\documentclass[letterpaper]{article} 
\usepackage{aaai23}  
\usepackage{times}  
\usepackage{helvet}  
\usepackage{courier}  
\usepackage[hyphens]{url}  
\usepackage{graphicx} 
\usepackage{natbib}  
\usepackage{caption} 
\usepackage{algorithm}
\usepackage[noend]{algpseudocode}
\usepackage{newfloat}
\usepackage{listings}
\DeclareCaptionStyle{ruled}{labelfont=normalfont,labelsep=colon,strut=off} 
\floatstyle{ruled}
\newfloat{listing}{tb}{lst}{}
\floatname{listing}{Listing}
\usepackage{mathrsfs}
\usepackage{amsmath}
\DeclareMathOperator*{\argmax}{arg\,max}
\DeclareMathOperator*{\argmin}{arg\,min}
\DeclareMathOperator*{\concat}{%
    \mathchoice%
        {\Big\Vert}%
        {\big\Vert}%
        {\Vert}%
        {\Vert}%
}
\usepackage{amsthm}
\theoremstyle{definition}
\newtheorem{definition}{Definition}[section]
\usepackage{sidecap}
\usepackage{wrapfig}
\usepackage{subfig}
\usepackage{amsfonts}           

\title{Global Concept-Based Interpretability for Graph Neural Networks via Neuron Analysis}
\author{
    Han Xuanyuan\textsuperscript{\rm 1},
    Pietro Barbiero\textsuperscript{\rm 1},
    Dobrik Georgiev\textsuperscript{\rm 1},
    Lucie Charlotte Magister\textsuperscript{\rm 1},
    Pietro Li\`o\textsuperscript{\rm 1}
}
\affiliations{
    \textsuperscript{\rm 1}University of Cambridge\\
    hx263@cantab.ac.uk \{pb737, dgg30, lcm67, pl219\}@cam.ac.uk 
}

\usepackage{bibentry}

\begin{document}

\maketitle

\begin{abstract}
Graph neural networks (GNNs) are highly effective on a variety of graph-related tasks; however, they lack interpretability and transparency. Current explainability approaches are typically local and treat GNNs as black-boxes. They do not look inside the model, inhibiting human trust in the model and explanations. Motivated by the ability of neurons to detect high-level semantic concepts in vision models, we perform a novel analysis on the behaviour of individual GNN neurons to answer questions about GNN interpretability. We propose a novel approach for producing global explanations for GNNs using neuron-level concepts to enable practitioners to have a high-level view of the model. Specifically, (i) to the best of our knowledge, this is the first work which shows that GNN neurons act as concept detectors and have strong alignment with concepts formulated as logical compositions of node degree and neighbourhood properties; (ii) we quantitatively assess the importance of detected concepts, and identify a trade-off between training duration and neuron-level interpretability; (iii) we demonstrate that our global explainability approach has advantages over the current state-of-the-art -- we can disentangle the explanation into individual interpretable concepts backed by logical descriptions, which reduces potential for bias and improves user-friendliness. 
\end{abstract}

\section{Introduction}

Graph neural networks (GNNs) have shown promising results in a variety of tasks where graph-structured data is prevalent \cite{WIEDER20201, SanchezGonzalez2020LearningTS, Wu2021GraphNN}. Like most neural methods, GNNs lack explainability and interpretability. Their complex decision-making is effectively a black-box process to a human observer \cite{newsurvey}, which inhibits human trust in critical applications such as healthcare \cite{xaihealthcare, xaiinfinance, xaiinlaw}. Although there has been significant research on the behaviour of individual neurons of convolutional neural networks (CNNs) and recurrent neural networks (RNNs) \cite{netdissect,Dalvi2019WhatIO,samek2019explainable}, existing approaches for explaining GNNs tend to treat the model as a black-box and only consider input feature attribution \cite{newsurvey}, ignoring the potential of concepts as a way of interpreting models, which have been shown to better align with human intuition due to the ability to present an explanation in terms of interpretable units \cite{Kim2018InterpretabilityBF, Koh2020ConceptBM,Ghorbani2019TowardsAC}. We perform an  investigation into the presence of concepts in GNNs, and propose a new approach for global GNN explainability. We make the following contributions:
\begin{enumerate}
    \item 
    We show that GNN neurons have alignment with concepts formulated as logical compositions of node and neighbourhood properties, and act as concept detectors which detect \textit{interpretable units of information} such as chemical substructures and social network motifs.  
    \item
    We propose new evaluation metrics adapted for GNN neurons. We quantitatively analyse the relationship between neuron importance and interpretability, and also show that there is a trade-off between model training and interpretability: GNNs trained for longer become harder to interpret, even if model performance plateaus.
    \item
    We compare our global GNN explanations against the existing state-of-the-art approach and show that our method possesses advantages for GNN users, including being concept-based, providing explanations backed by logical descriptions, and being inherently noise-robust.
\end{enumerate}

\section{Background}
\begin{figure*}[!htb]
\centering
\includegraphics[width=.7\textwidth]{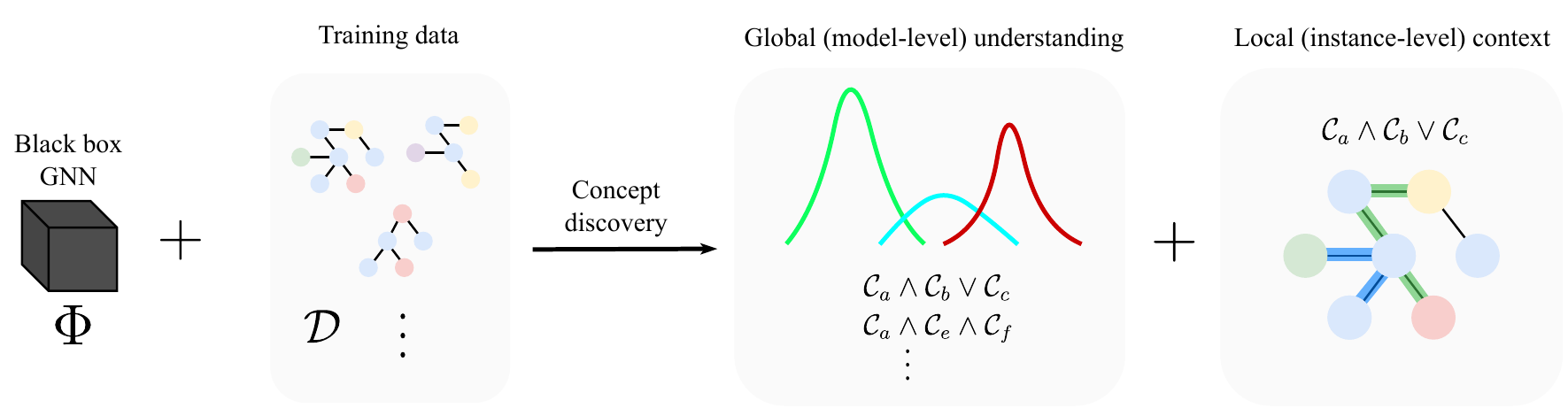}
\caption{The proposed framework. Given a black-box GNN $\Phi$ and training set of graphs $\mathcal{D}$, we (i) discover interpretable neurons in $\Phi$ that act as concept detectors by performing a search over $\mathcal{D}$; (ii) utilise the discovered concepts to produce global explanations; (iii) adapt the concepts to specific local instances from $\mathcal{D}$ to produce visualisations to aid the practitioner.}
\label{overview}
\end{figure*}
Explainablility for GNNs is a relatively new direction. Most approaches focus on local post-hoc explainability using input attribution via methods such as mutual information maximisation. For example, \citet{gnnexplainer} proposed GNNExplainer. Many other works focus on providing local explanations which are more robust and faithful \cite{newsurvey,pgexplainer,Vu2020PGMExplainerPG}. On the other hand, a global or \textit{model-level} explanation tells the practitioner how the model behaves as a whole.
Model-level (or global) approaches are less explored for GNNs. The current state-of-the-art model-level approach is XGNN, which uses reinforcement learning to generate a graph that maximises the prediction of a certain class \cite{xgnn,newsurvey}. Unfortunately, since their form of explanation is a graph generated using a black-box neural network, the meaning of the graph is not always intelligible and leaves interpretation to the user, which can lead to human bias \cite{himmelhuber2021new}. However, a recent trend in explainability is to use \textit{concepts}, which have been shown to make explanations more plausible and less dependent on human interpretation due to their ability to express a model's decision making in terms of high-level interpretable units of information \cite{Kim2018InterpretabilityBF, Koh2020ConceptBM,Ghorbani2019TowardsAC}. We assume the definition of concepts proposed by \citet{Ghorbani2019TowardsAC}, i.e. a concept should ideally satisfy the properties of \textit{meaningfulness}, \textit{coherency} and \textit{importance}. Concepts have been applied to GNNs by \citet{gcexplainer}, who treat a concept as a $k$-means cluster in the activation space, and also to the task of algorithmic reasoning via a concept bottleneck layer \citep{Koh2020ConceptBM,acer}. Moreover, in the vision and language domains much work on probing individual model neurons has been done \cite{netdissect,Dalvi2019WhatIO,muandreas}, and it has been found that individual neurons can behave as \textit{concept detectors}. The potential of such an approach to address GNN interpretability has been unexplored.

\subsection{Graph neural networks}
We define a graph $G=(E,V)$ as a set of edges $E$ and nodes $V$. A GNN model is a function $\Phi(\mathbf{X}, \mathbf{A})$ such that $\mathbf{X} \in \mathbb{R}^{D \times |V|}$ is a node feature matrix, where each column $\mathbf{x}_i$ is the $D$-dimensional feature vector of node $i$, and $\mathbf{A} \in \left  \{0,1\right\}^{|V| \times |V|}$ is an adjacency matrix which represents the edges $E$. Although many GNN architectures exist, most can be categorised under \textit{message passing} GNNs \cite{messagepassing}. At each layer of message passing, node representations $\mathbf{H} \in \mathbb{R}^{D \times |V|}$ are updated using message passing function $f$, i.e. $\mathbf{H}^{(l+1)}=f(\mathbf{H}^{(l)},\mathbf{A})$. We present our method assuming that the GNN follows this general framework. Furthermore, although there exists many ways of initialising the node representations, we assume that $\mathbf{H}^{(0)}=\mathbf{X}$. We shall refer to $\mathbf{H}^{(l)}$ as the activations at the $l$-th layer, and $\mathbf{H}^{(l)}[i,:]$ as the $i$-th \textit{neuron} at that layer\footnote{We use slicing notation. $\mathbf{H}[i, :]$ indicates the $i$-th row of $\mathbf{H}$, and  $\mathbf{H}[:, j]$ the $j$-th column.}.

\section{Methodology}

In this section we outline the proposed method. A visual overview is shown in Figure \ref{overview}. 

\subsection{Concept discovery}
\label{sec:discovery}
With inspiration from the concept extraction method by \citet{muandreas} designed for vision and language models, we propose to perform a compositional concept search over the activation space of a GNN.
Formally, letting $\mathcal{G}$ be the space of input graphs, we treat concepts as functions $C : \mathcal{G} \rightarrow \{0,1\}^{|V|}$ which outputs a boolean value indicating whether each node is part of the concept, i.e. binary concept \textit{masks}. We define a concept $C$ to be \textit{interpretable} if there exists a natural language description of $C$ such that a practitioner is able to accurately predict the output of $C$ given any graph $G \in \mathcal{G}$ from solely knowing this description. Hence, concepts such as \textit{nodes that have exactly one neighbour} would be interpretable. Furthermore, we refer to the \textit{space} of concepts as a set $\mathcal{C}$ which contains all concepts that are interpretable.

Our goal is to determine whether neurons are concept detectors. For a graph $G$, let $V_G$ be the nodes in $G$, $C(G)[i]$ indicate the mask value for the $i$-th node, and $\mathbf{H}_G^{(l)}$ be the activations of the model taking $G$ as input. Note that $\mathbf{H}_G^{(l)}[k,i]$ indicates the $k$-th neuron activation of the $i$-th node. We formally define a neuron $k$ to be a concept detector if 
\begin{equation}
\exists C. \forall G \in \mathcal{G}. \left[ \forall i \in V_G . \left[ C(G)[i] \approx \tau_k \left ( \mathbf{H}_G^{(l)}[k,i] \right ) \right] \right].
\end{equation}
where $\tau_k : \mathbb{R}^m \rightarrow [0,1]^m$ is an element-wise thresholding function specific for neuron $k$. This formulation assumes that there are concepts with near-perfect alignment with the activations, which in practice is unrealistic. 
Therefore, we relax the definition and consider whether there are concepts that approximately resemble the behaviour of a neuron $k$:
\begin{equation}
\exists C.\ \mathbb{E}_{G \sim \mathcal{G}} \left [ \text{Div} \left (C(G), \mathbf{H}_G^{(l)}[k,:], \tau_k \right ) \right ] \approx 0.
\end{equation}
Here, $\text{Div}$ is some measure of divergence between the concept mask $C(G)$ and the activations thresholded using $\tau_k$. Following previous works on image models \cite{muandreas,netdissect} that have used intersection over union (IOU) scores, we choose a similar metric
\begin{equation}
\text{Div}\left (\mathbf{a},\mathbf{b},\tau \right )=-\frac{\sum_i (\mathbf{a} \cap \tau(\mathbf{b}))_i}{\sum_i (\mathbf{a} \cup \tau(\mathbf{b}))_i}\cdot\frac{\mathbf{a} \cdot \tau(\mathbf{b})}{\sum_i b_i},
\end{equation}
which is the IOU score scaled by a factor representing how much of the activation signal is incorporated into the concept. Note that $\mathbf{a} \cdot \tau (\mathbf{b})$ represents the total activations within the concept mask, and $\sum b_i$ is the sum of all activations in the graph. 
Following previous work \cite{muandreas,netdissect}, we apply dynamic thresholding to obtain $\tau$. The search objective becomes the process of finding a concept that maximises a score $f(C,k)$ for neuron $k$. 
\begin{multline}
\argmax_{C} f(C,k) \ \text{s.t.} \ f(C,k)= \\ \min_{\tau_k} {|\mathcal{D}|}^{-1}\sum_{G\in\mathcal{D}} \text{Div}\left (C(G), \mathbf{H}_G^{(l)}[k,:], \tau_k \right)
\end{multline}
in which we search through the concept space $\mathcal{C}$. This is tractable if $|\mathcal{C}|$ is small. However, following  \citet{muandreas} we desire to search for \textit{compositional} concepts, i.e. those which are a logical composition over base concepts. Formally, we define a compositional concept of length $k$ as an expression of the form
\begin{equation}
C_1 \oplus_1 C_2\oplus_2 \ldots \oplus_3 C_k,
\end{equation}
where $\{C_i\}_{i=1:k} \subseteq \mathcal{C} \cup \{\neg C \mid C \in \mathcal{C}\}$ and $\{\oplus_i\}_{i=1:k} \subseteq \{\wedge,\vee\}$, which is intractable for large $k$. 
A beam search is performed over the space of compositional concepts using the concept divergence as a priority metric to prune the less desirable concepts. The full algorithm and implementation details are provided in Appendix \ref{appendix:conceptextraction}\footnote{The full paper along with the appendix can be found on ArXiv: \url{https://arxiv.org/abs/2208.10609}. The code is available at \url{https://github.com/xuyhan/gnn-dissect}.}. Note that in practice, we use a GPU-based implementation to perform this search on all neurons in parallel.

\subsection{Concept-based model-level explanations}
\label{sec:global}
Given GNN $\Phi$ and target class $y$, we propose to provide model-level explanations using the concept-extraction framework. 
We follow the concept-bottleneck paradigm \cite{Koh2020ConceptBM,lens} of separating the top-level of $\Phi$ (the section after the final message passing round) from the rest of the model. Denoting this top level with $\Phi_T$, we propose to produce an explanation for $\Phi_T$ as a set of neurons $S_\text{glob}$ that globally explains the behaviour of $\Phi$ for class $y$. Then, we align the explanation with concepts by finding $\mathcal{E}_\text{glob}=\{ \argmax_{C \in \mathcal{C}} f(k, C) \mid k \in S_\text{glob} \}$, which is the set of highest scoring concepts for neurons in $\mathcal{S}_\text{glob}$. For models where $\Phi_T$ is a single perceptron, we propose to select the neurons with positive weight contributions for class $y$. 

Due to the importance of visualisation for interpretability \cite{samek2019explainable}, for each concept in $\mathcal{E}_\text{glob}$ we also produce a visualisation to accompany the concept using \emph{activation maps} which are a common approach for vision models \citep{gradcam,cam}. However, a clear distinction is that our activation maps are concept-based and operate over graphs (rather than images) -- we therefore refer to them as \emph{graph concept activation maps}.
\begin{definition}[Graph concept activation map]
Given GNN $\Phi$ with $L$ layers, graph $G=(V,E)$, target explanation class $y$ and neuron $k$, the concept activation map for $(\Phi, G, y, k)$ is $(C_k, \eta_V^k,\eta_E^k)$ where $C_k$ is the best-matching concept for neuron $k$. Also,
$\eta_V^k \in \mathbb{R}^{|V|}$ and $\eta_E^k \in  \mathbb{R}^{|V|\times |V|}$ are importance masks over the nodes and edges respectively. Inspired by activation maps on CNNs \cite{gradcam,cam}, we take $\eta_V=\frac{\partial \Phi(G)_y}{\partial n_G^k} \cdot \mathbf{H}_G^{(L)}[k,:]$ where $\Phi(G)_y$ is the output logit for class $y$, and $n_G^k=\oplus_{j=1}^{|V|} \mathbf{H}_G^{(L)}[k,j]$ is the pooled value for neuron $k$ in which $\oplus$ is a global pooling operator. Now for $\eta_E^k$ we aggregate the scaled node activations. Denoting $\boldsymbol\eta_V^k=(\eta_V^k, \eta_V^k, \ldots, \eta_V^k) \in \mathbb{R}^{|V|\times|V|}$, we compute $\eta_E^k=(\boldsymbol\eta_V^k   + (\boldsymbol\eta_V^k)^\top) \odot A$ where $\odot$ is the Hadamard product and $A$ is the adjacency matrix of $G$. 
\end{definition}
Now, for each neuron $k \in \mathcal{S}_\text{glob}$  we choose $G=\argmax_{G\in\mathcal{D}} \text{max}_j \mathbf{H}_G[j, i]$ as the graph that has the highest presence of neuron $k$, and produce the concept activation map for $(\Phi,G,y,k)$. 

\paragraph{Contextualising global explanations using instances} The proposed method is able to extract concepts on a model-level. This is accompanied by a visualisation on an instance-level to aid the explanation. As there is growing interest in relating global explanations to local instances \cite{lundberg2020local}, we introduce metrics to assess how important a concept is in the model's decision:
\begin{itemize}
    \item \textit{Absolute contribution} (ABS): the value of the neuron's weight connection to the class output.
    \item \textit{Entropy score} (ENT): we propose a mutual-information maximisation approach inspired by GNNExplainer \cite{gnnexplainer} to find the set of neurons $S_\text{loc}$ which best explains a local prediction, whereby we optimise the objective
    \begin{equation}
        S_\text{loc}=\argmin_{S} \ - \log P_\Phi(Y=y \mid \mathcal{X}=\sigma(\eta_N)),
    \end{equation}
    in which $P_\Phi$ is the logit for class $y$, $\mathcal{X}$ is the set of neurons utilised by the model, and $\eta_N$ is a learned mask over the neurons. The full derivation is shown in Appendix \ref{appendix:entropy}. The entropy importance of neuron $k$ is mask value $\eta_N[k]$.
\end{itemize} 

\begin{figure*}[htb!]
\centering
\includegraphics[width=.95\textwidth]{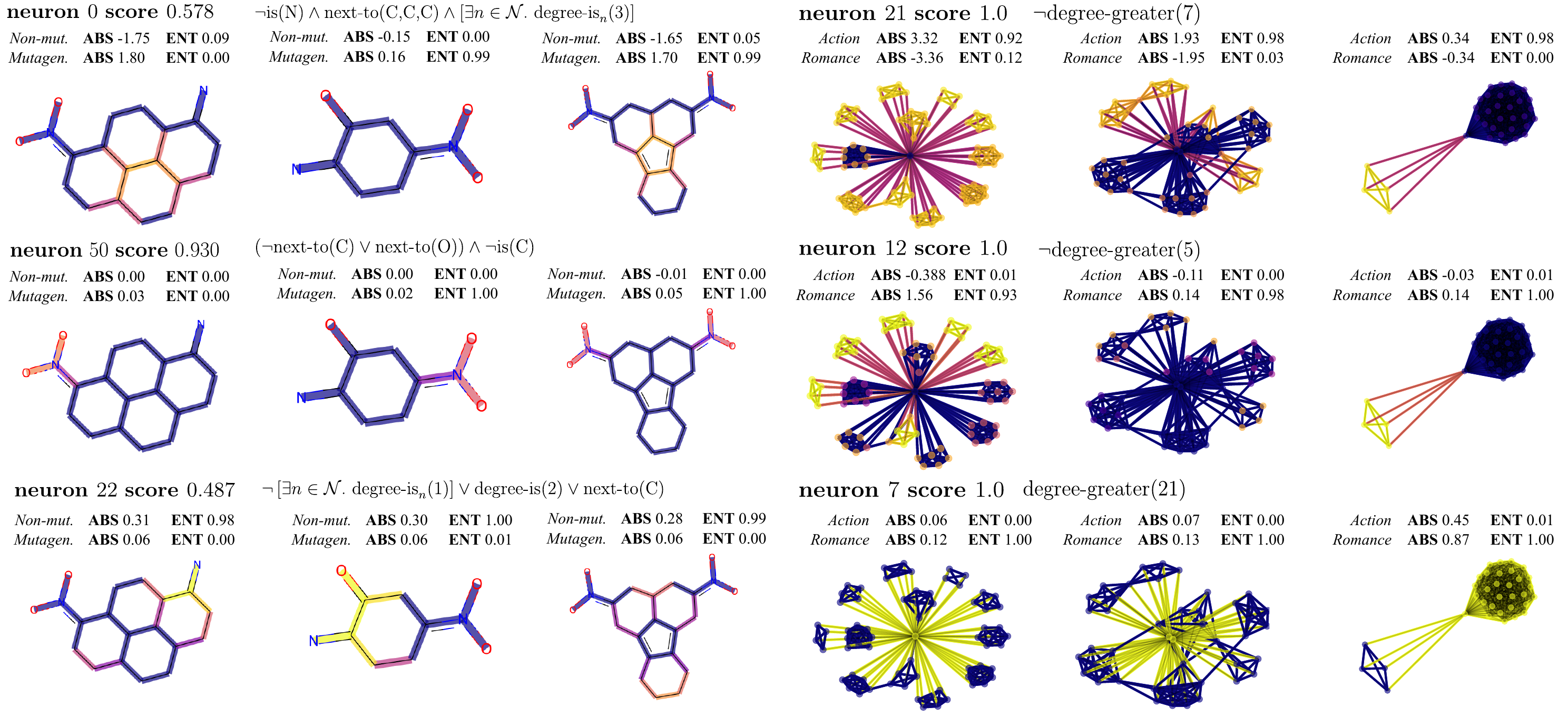}
\caption[MUTAG Concepts.]{Concepts in MUTAG (left) and IMDB-B (right). Brighter colours indicate higher importance. Each row is thresholded using the  threshold $\tau_k$.}
\label{mutagimdb}
\end{figure*}
\section{What concepts do GNNs learn?}
\label{sec:whatconcepts}

In this section we use the concept extraction approach to investigate the types of concepts that arise in GNNs, and demonstrate that they align with human intuition.

\paragraph{Datasets and models} We perform experiments on four popular real-world benchmark datasets for binary graph classification: MUTAG \cite{mutag}, Reddit-Binary \cite{deepgraphkernels}, PROTEINS \cite{proteins} and IMDB-Binary \cite{deepgraphkernels}. Following the existing trend in GNN explainability, we choose to train models which fit under the convolutional flavour of GNNs. For each graph classification task we train a model with three layers of the Graph Isomorphism Network (GIN) architecture \cite{gin}. The exception is Reddit-Binary, where we use graph convolutional network (GCN) layers \cite{gcn} as we observed training to be more stable given the large graph sizes. The exact descriptions of the models and their training parameters are shown in Appendix \ref{appendix:models}. The same concept discovery procedure is run for all datasets, with a different set of base concepts for each. These were determined based on the nature of the task. For example in MUTAG, the nodes can be one of seven types, and it is natural to include these as base concepts.  
\subsection{Concepts for each model}
We apply our GNN concept discovery framework on the models trained on the benchmark datasets, using a set of base concepts for each task to align with the human-in-the-loop nature of explainability \cite{SIMKUTE2021100017}. We choose \textit{simple} concepts that non-domain experts are able to understand, such as \textit{next-to(X)} (being next to a node with label $X$), and \textit{is(X)} (having label $X$). The full set of base concepts for each task and their descriptions is shown in Appendix \ref{appendix:base}. We find that interpretable concepts \textit{are} detected by the neurons of a GNN, and they vary across the different tasks. Using concept activation maps we produce visual summaries of the most intepretable concepts. The concept contribution scores for each example graph using \textit{absolute contribution} and \textit{entropy score} are provided. 

\paragraph{MUTAG} We find that concepts in MUTAG correspond to known \textit{functional groups}. As shown in Figure \ref{mutagimdb}, we find $\text{NO}_2$ detector concepts and carbon ring concepts, which corroborate with the findings of \citet{gnnexplainer}. The $\text{NO}_2$ concept is in fact a known mutagen which has been shown to be a strong indicator of mutagenicity \cite{toxicophore}, and the presence of benzene carbon rings is an indicator of aromaticity which together with $\text{NO}_2$ is known to indicate mutagenicity. In the examples in Figure \ref{mutagimdb}, \textbf{neuron 0} is a carbon ring detector and \textbf{neuron 50} is an $\text{NO}_2$ detector. We observe that carbon ring concepts typically have much higher absolute contribution to the final prediction. Unlike prior work, we additionally observe concepts which contribute to the prediction of non-mutagenicity. In particular we observe concepts that are highly concentrated on parts of compounds that hang off carbon rings but are not $
\text{NO}_2$ substructures - for instance, \textbf{neuron 22} appears to detect any part of the input which is not part of a carbon ring or an $\text{NO}_2$ substructure. Other concepts appear to trigger for only certain types of atoms hanging off carbon rings, e.g. \textbf{neuron 28} (not shown) detects atoms that are not fluorine nor chlorine, and have exactly one carbon neighbour. 
\begin{figure}[!htb]
\centering
\includegraphics[width=.45\textwidth]{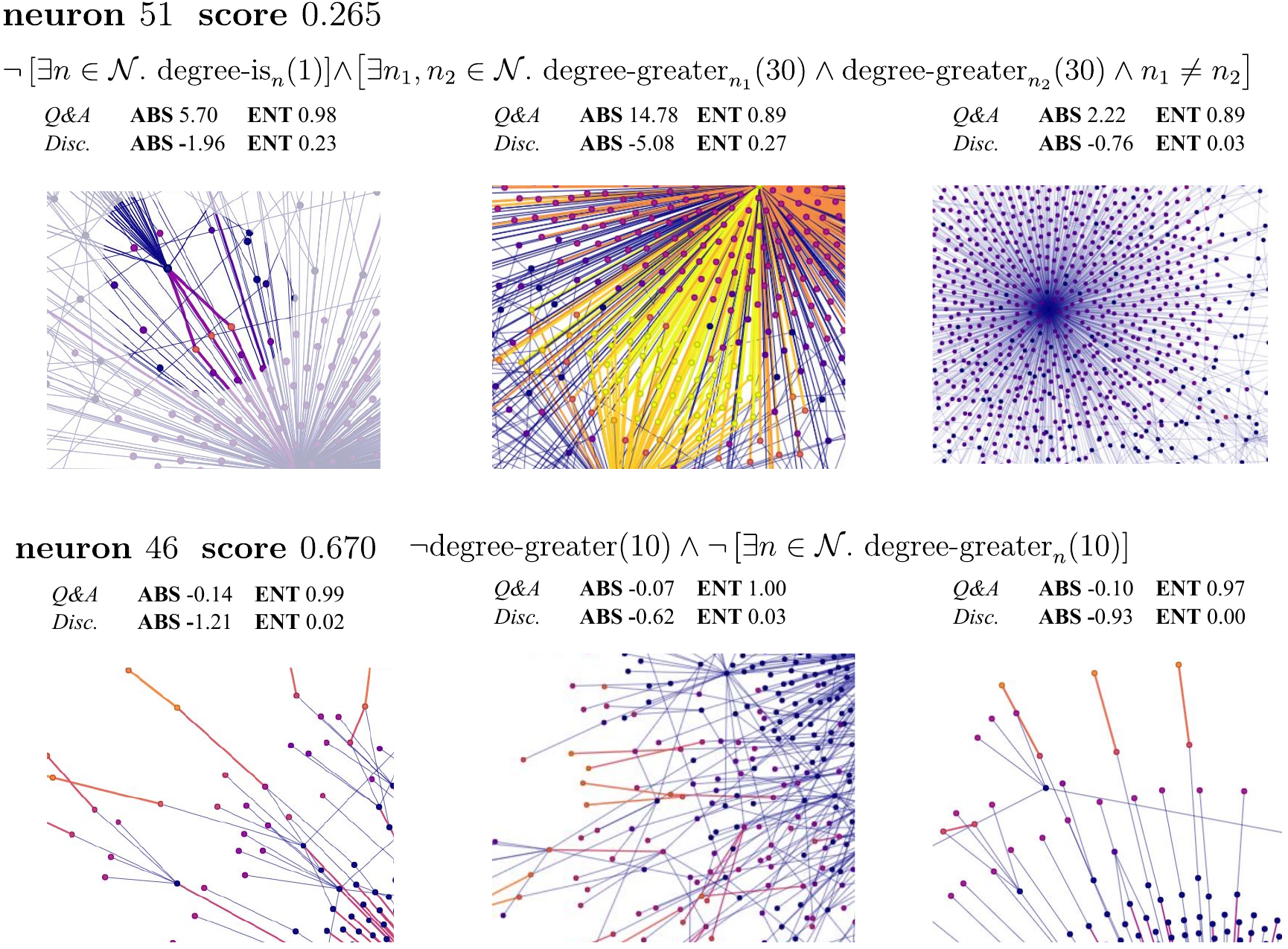}
\caption[Reddit-Binary Concepts.]{Reddit-Binary Concepts. Node colours represent activation values for the concept, with brighter colours indicating higher values. For graphs where the concept is hard to see by eye, we have highlighted the concept area by changing the opacity of the image. }
\label{redditvis}
\vspace{-1.5em}
\end{figure}
\paragraph{Reddit-Binary} Examples of concepts are shown in Figure \ref{redditvis}. In Reddit-Binary, graphs typically have one or two nodes with high degree, corresponding to the highly voted comments that receive the most attention. We observe the emergence of the `few-to-many' subgraphs, resembling a small number of experts replying to several questions, as was mentioned by \citet{gnnexplainer}, and is a strong indication that the graph is of the Q\&A class. An example of such a concept detector is \textbf{neuron 51}. We also see neurons which appear to detect the nodes that are more than a single hop from a high-degree node, representing users at the end of a long comment chain, e.g. \textbf{neuron 46}. We observe that these concepts penalise the discussion class in terms of absolute contribution. One possible explanation is that discussion threads typically involve the same users commenting several times in the same chain, which would cause the chain to collapse in the graph (since nodes represent users, not comments). We additionally find neurons that appear to trigger when their degree is above a certain threshold, such as \textbf{neuron 20} (not shown). Intuitively, a user with lots replies is a strong indication of Q\&A, since the host of the Q\&A will likely receive a large number of follow-up questions.

\paragraph{IMDB-Binary} Graphs typically have one or two `popular' nodes that have high degree, and have multiple smaller communities attached to them. As shown in Figure \ref{mutagimdb}, we observe that almost all active neurons are sensitive to the \textit{degree} of a node, with some neurons only activating if the degree exceeds (or is less than) a certain amount. We find that \textbf{neuron 21} is activated for the small communities that are attached to the popular node, and is accurately summarised as detecting if the degree is no greater than 7. It acts as a strong indicator of the genre being \textit{action}. Interestingly, \textbf{neuron 12} behaves similarly  - activating for nodes with degree no more than 5 - but is a strong indicator for the opposite genre \textit{romance}. This suggests the GNN is learning to discriminate genres based on subtle differences in the size of the small communities. There exists high-degree concepts, e.g. \textbf{neuron 7} which are indicative of the romance class. Moreover, we are surprised by the high interpretability score of these neurons, with many having a perfect score of $1$, i.e. the concepts learnt by the GNN can be summarised accurately using only degree information.  

\paragraph{PROTEINS} We provide additional findings with the PROTEINS dataset in Appendix \ref{appendix:proteins}.

\subsection{Discussion}

Overall we observe that a range of concepts are being detected by GNN neurons. We make three observations.

\begin{enumerate}
\item
We first observe that the extracted concepts are more interpretable than prior work for vision and language models. This is attributed to the fact that GNNs tend to be much shallower than CNNs and RNNs, and the ones used in the experiments have at most three layers. This is perhaps a hopeful sign from the perspective of GNN interpretability and the desirability for concepts to be extracted without resorting to manual labelling. 
\item
Secondly, we see that not all concepts are useful for the model. There exist neurons which detect concepts, but have low contribution to predictions. In addition, there exists neurons that the model believes to be useful for \textit{both} classes and therefore do not act as useful discriminators for the model. For example, in IMDB-Binary, there are neurons that detect the presence of a node with large degree which have positive weights for both \textit{romance} and \textit{action} genres. This suggests that by manually removing or suppressing certain concepts we may be able to reduce the number of parameters whilst preserving performance, which serves as a form of \textit{model pruning}. We leave this idea open for future investigation.
\item
Moreover, there are often neurons that detect similar concepts, and appear to be redundant in nature. For instance, the MUTAG model contains at least 3 neurons that act more or less as $\text{NO}_2$ detectors, but their associated concepts are marginally different. This can be attributed to the nature of SGD-based algorithms used to train GNNs - there is no explicit constraint that enforces each neuron to behave differently. We note that despite there being many neurons detecting similar concepts, their activation magnitudes can still vary greatly. This suggests the method can be potentially used to identify neurons that are redundant in the information they learn.
\end{enumerate}

\subsection{Are interpretable concepts more important?}

A natural question to ask is whether the concepts identified are beneficial to the model making correct predictions. 
We define a neuron's \textit{interpretability} as the score of its best matching concept. Existing methods of determining neuron importance do not necessarily work for GNNs as they do for CNNs. For example, we found that \textit{erasure} \cite{li2016understanding} is a poor metric for determining neuron importance due to the high presence of \textit{redundant neurons} learnt by GNNs. 
We propose two alternative metrics.

\textbf{Neuron importance}. For a model with $r$ class outputs, we measure the importance of neuron $k$ in making a single prediction as the variance of individual logit contributions. Recall that for input $G=(E,V)$, $n_G^k=\oplus_{j=1}^{|V|} \mathbf{H}_G^{(L)}[k,j]$ is the pooled value of neuron $k$. Aiming to measure the neuron's ability to discriminate between classes, rather than its absolute contribution, we consider the variance of individual logit contributions, i.e. $\text{Var}(w^1_kn^k_G, \ldots, w^r_kn^k_G)$ where $w^j_k$ is the weight connecting $n^k_G$ to the $j$-th logit. The \textit{importance} score is this value averaged over all graphs in the dataset:
\begin{equation}
\text{importance}(k)=\frac{1}{|\mathcal{D}|}\sum_{G \in \mathcal{D}} \text{Var}(w^1_kn^k_G, \ldots, w^r_kn^k_G).
\end{equation}
\textbf{Neuron correctness}. Following \citep{muandreas} we also measure how much each neuron contributes to the model being \textit{accurate}. They chose to measure the frequency of correct predictions when each neuron is activated\footnote{While this yielded good results for CNNs, it does not transfer well to GNNs due to the redundant nature of neurons}. Let $y_G$ be the predicted logits for input $G$ and $\hat{y}_G$ be the ground truth. Treating $n^k_G$, $y_G$, and $\hat{y}_G$ as random variables sampled across $G \in \mathcal{D}$, we choose to compute a correctness score for each neuron $k$ defined as \begin{equation}
\text{correctness}(k)=\lvert\text{Corr}(n^k_G, -\mathcal{L}(y_G, \hat{y}_G))\rvert,
\end{equation}
in which we measure the Pearson product-moment correlation between the neuron's magnitude with cross-entropy loss $\mathcal{L}$.
\begin{figure}[!h]
\centering
\subfloat[\centering MUTAG]{{\includegraphics[height=5cm]{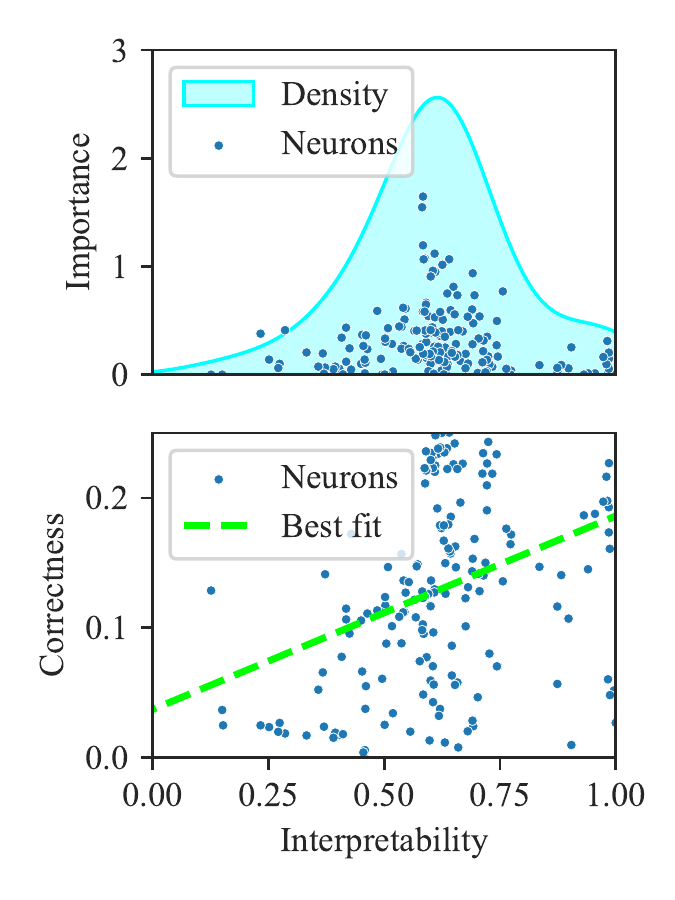} }}%
\subfloat[\centering Reddit-B]{{\includegraphics[height=5cm]{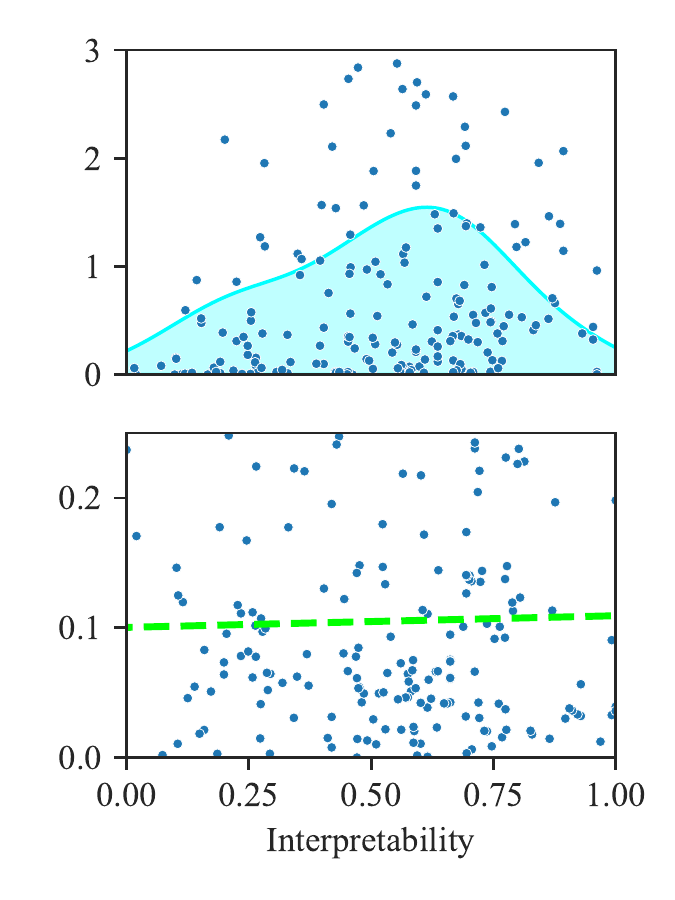} }}%
\caption{Investigating the interpretability of neurons and their importance. Kernel density estimates \cite{kde} are shown in the top row to indicate the relative spread of neurons across different interpretability scores. Dashed lines are lines-of-best-fit to indicate trend. }
\vspace{-1em}
\label{neuronanalysis}
\end{figure}

We investigate the relationship between interpretability and the aforementioned metrics, by training ten models for each dataset across different train-test splits\footnote{IMDB-Binary is excluded since almost all neurons are highly interpretable}. We show the results of the analysis in Figure \ref{neuronanalysis}. First, for the MUTAG and Reddit-Binary tasks, the distribution of neuron interpretability scores appears to resemble a bell curve, where more neurons have scores closer to 0.5 than either extreme. In terms of the relationship with neuron importance, we observe that the most important neurons are situated around the centre of the interpretability spectrum. This suggests that the neurons that are most important are also more difficult to summarise using a set of interpretable rules. Nevertheless, we do not observe a negative correlation between neuron importance and interpretability, and the most important neurons typically have around 0.5 interpretability, which we emperically observe to be sufficient in terms of explanation quality. In terms of contribution to model accuracy, we observe that neuron interpretability is positively correlated with having accurate predictions for MUTAG, and having slight positive correlations for Reddit-Binary. 

\subsection{How do concepts vary across models?}

Works on interpreting deep models with concepts typically focus on well-trained models that generalise well. They also focus on the last few hidden layers since high-level concepts are more likely to emerge there. This motivates us to ask two follow-up questions:
\begin{enumerate}
    \item Do models with only a few epochs of training exhibit different concepts than models that are trained for extended numbers of epochs?
    \item Do different types of concepts emerge at different depths of the model?
\end{enumerate}


We perform additional experiments on the MUTAG and Reddit-Binary datasets in an attempt to answer these questions and provide some preliminary steps for future investigation. 

\begin{figure}[!ht]
\centering
\subfloat[\centering MUTAG]{{\includegraphics[height=5cm]{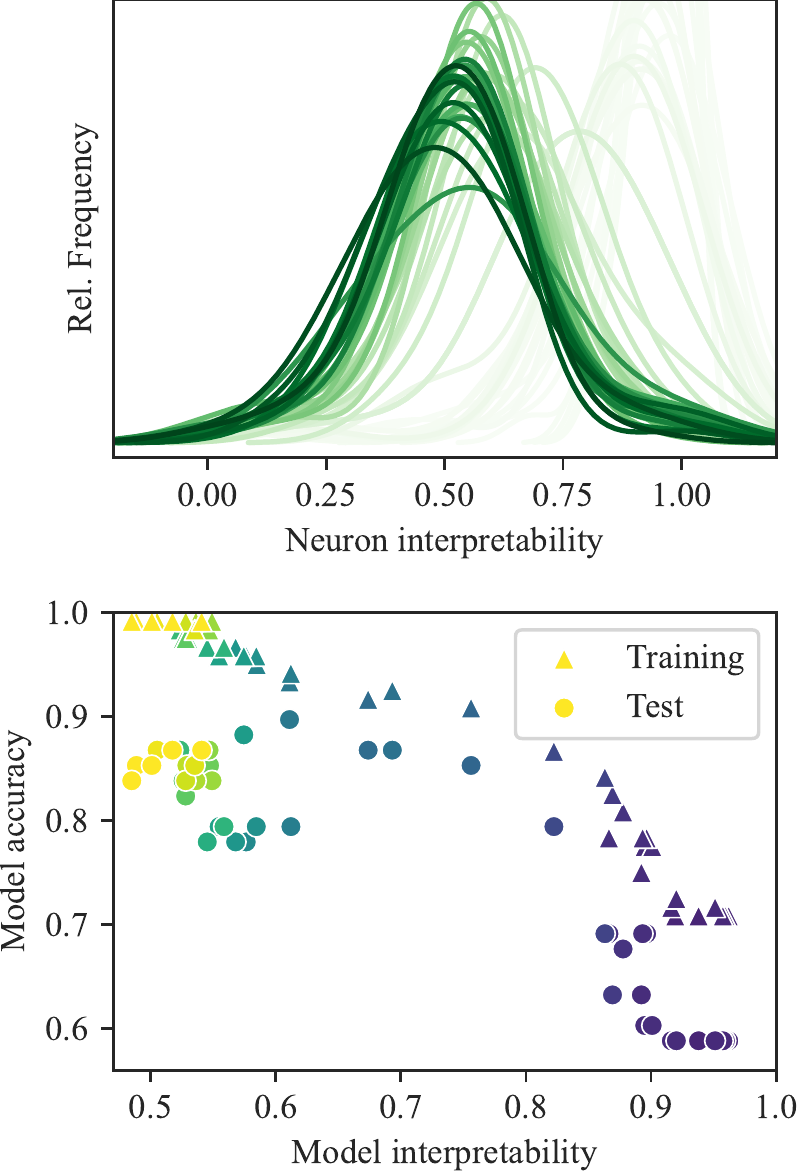} }}%
\subfloat[\centering Reddit-B]{{\includegraphics[height=5cm]{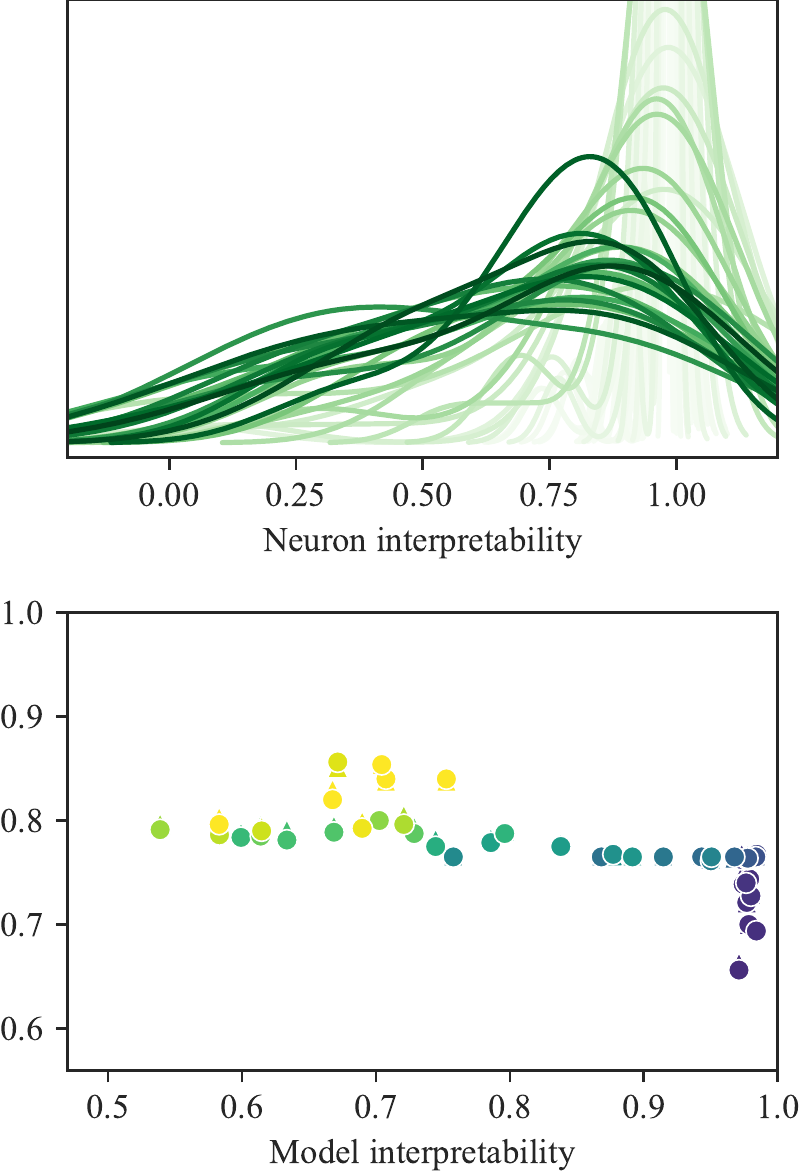} }}%
\caption{Training duration and model interpretability. The top row shows the distribution of neuron interpretability scores for models trained for varying epochs; darker curves correspond to more epochs. The lower row shows the relationship between the interpretability of a model and its accuracy; brighter points correspond to more epochs.}
\label{additional}
\end{figure}

\paragraph{Longer training reduces interpretability.} We train models using the same architectures as previously specified, but for varying numbers of epochs ranging from 1 to 400. For each model, we extract concepts from the final layer. Defining \textit{model interpretability} as the mean interpretability of its neurons, we observe the interpretability distributions for these models, as well as the relationship between model interpretability and accuracy. This is shown in Figure \ref{additional}. Training a model for longer periods causes concepts to be less interpretable, in general. Note that in some cases, there is a small difference in test set performance, despite a large difference in concept interpretability. For example, both the MUTAG models trained for 100 epochs and 400 epochs have the same test accuracy, but one has significantly more interpretable concepts. This phenomenon is more extreme on Reddit-Binary; we observe that for the first 50 epochs the model performance improves without much sacrifice in interpretability; however, later the interpretability drops significantly without much improvement in model performance. This suggests that there may be some general trade-off between interpretability and model accuracy, even when the same model architecture is used. Therefore, interpretability could be a factor to be considered during model training. A related theoretical issue is the \textit{information bottleneck} in deep learning \cite{shwartz2017opening}, which makes the observation that there is a trade-off between accuracy and the information flow in the latter layers of a model.
\begin{figure*}[!htb]
\centering
\subfloat[\centering Model-level explanation for \textit{mutagenic} on MUTAG]{{\includegraphics[height=4.5cm]{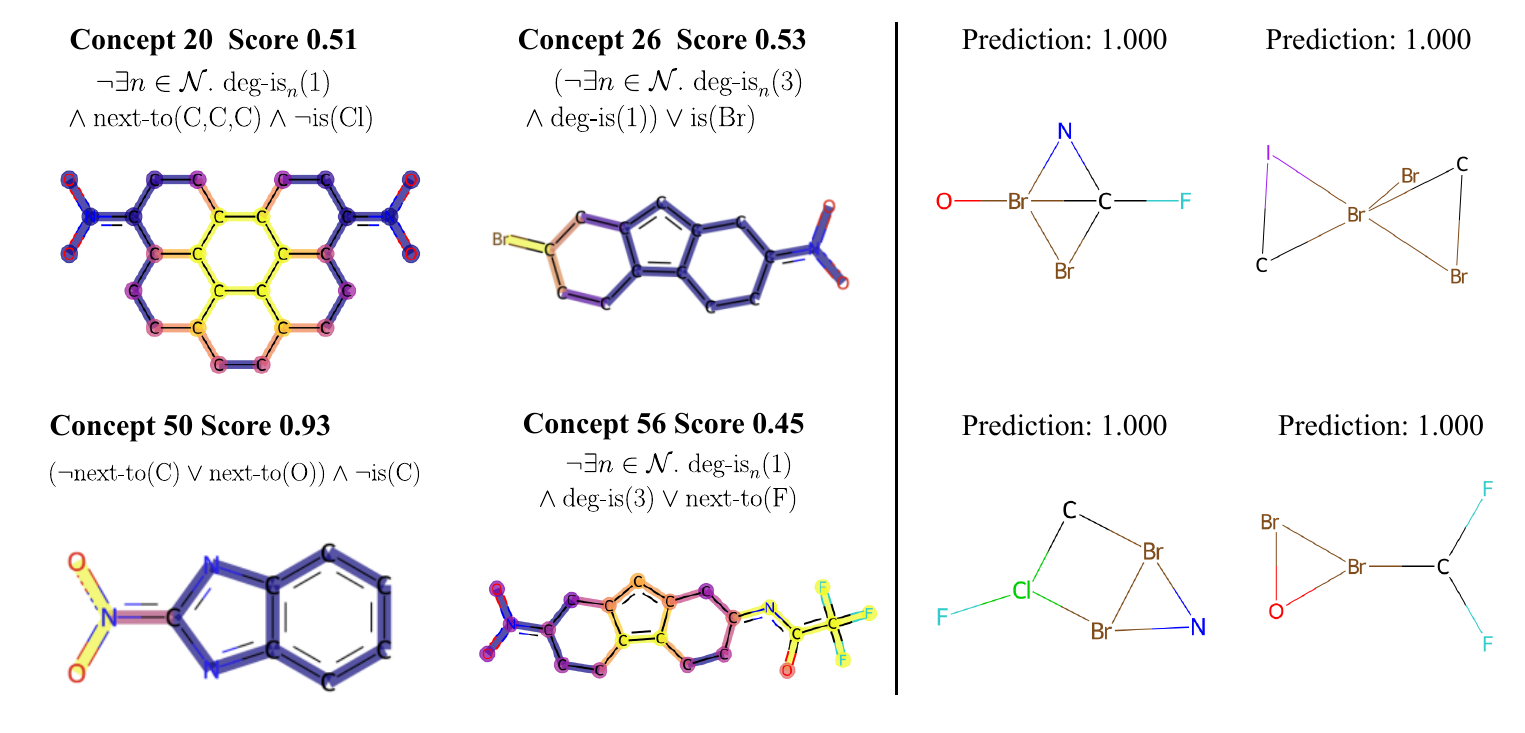} }}%
\subfloat[\centering Model-level explanation for \textit{romance} on IMDB-Binary]{{\includegraphics[height=4.5cm]{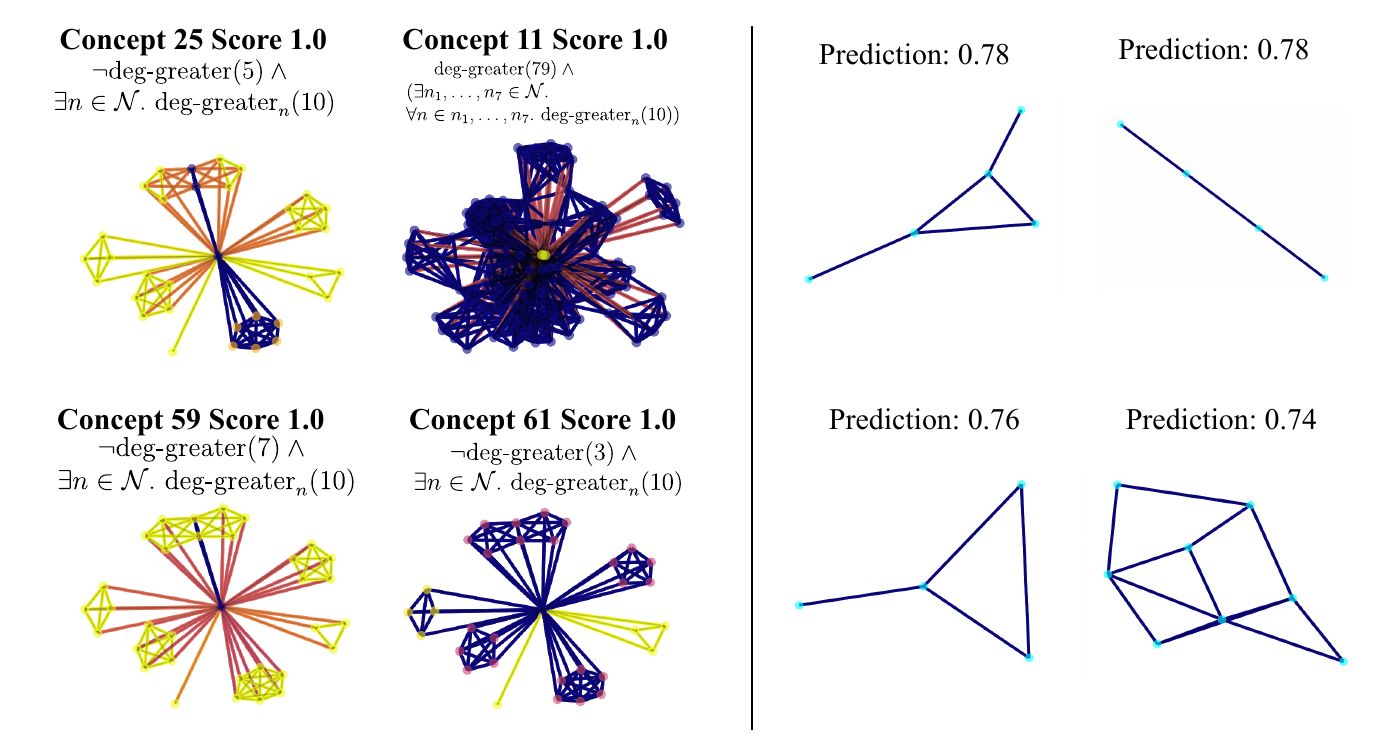} }}%
    
\caption[Comparison of class-specific model-level explanations against XGNN.]{Side-by-side comparison of class-specific model-level explanations against XGNN on MUTAG and IMDB-Binary. In each subfigure, our outputs are shown in the left; XGNN outputs are on the right along with the prediction values.}
\label{xgnncompare}
\end{figure*}

\paragraph{Concepts across layers.} We train additional models on the MUTAG and Reddit-Binary datasets, but using 10 layers of message passing. We probe the concepts after each layer and find that the mean neuron interpretability decreases further into the model. Refer to Appendix \ref{appendix:depth} for plots. The accuracy of these deeper GNNs is lower than the shallower versions, which is not surprising and likely due to \textit{over-smoothing} \cite{li2018deeper}. However, we also notice that the interpretability of the final layer is typically lower than those of the shallow models, suggesting that there is also a potential link between over-smoothing and neuron interpretability. 


\section{Model-level explanations}
\label{sec:expl}

    
In this section, we provide concept-based model-level explanations for the models investigated in Section \ref{sec:whatconcepts} using our approach described in Section \ref{sec:global}. We compare the explanations against the current state-of-the-art approach for model-level explanations for GNNs,  XGNN \cite{xgnn,newsurvey}, which produces class-specific global explanations via \textit{graph generation}. We compare against XGNN in terms of the model-level explanations produced by both methods, showcasing the concepts of our method against the generated graphs of XGNN. Comparisons on MUTAG and IMDB-Binary are shown in Figure \ref{xgnncompare}; additional comparisons on Reddit-Binary are provided in Appendix \ref{appendix:explanations}. 

\paragraph{Comparison} We want to emphasise that our approach offers desirable properties which XGNN is not capable of. In particular, it can be observed that:
\begin{itemize}
    \item
    Our approach is capable of separating the explanation into \textit{individual concepts}, which has been shown to align with human intuition 
    \cite{Koh2020ConceptBM}. XGNN is not concept-based and has no guarantee that each generated graph corresponds to a separate unit of decision making. 
    
    \item XGNN's generated graphs leave the human observer to draw a conclusion. Our approach, however, can provide logical descriptions of each concept together with an interpretability score. We argue that this reduces the potential for human bias compared to providing only a generated graph, which can have very different interpretations depending on the observer.
    \item
    XGNN often generates graphs that do not resemble members of the actual data distribution, making it difficult to contextualise the explanation in the problem setting. For example, we do not see the appearance of carbon-ring concepts for the MUTAG class \textit{mutagenic}. Similarly, we do not see the \textit{few-to-many} concept for Reddit-Binary. For IMDB-Binary, from the XGNN outputs we cannot infer that the class prediction is related to node degree. 
    \item
    We are able to contextualise our model-level explanations using local instances using the importance metrics proposed in \ref{sec:global}, which is not possible with XGNN.
    \item
    Our approach avoids the `explaining a black-box via another black-box' issue \cite{Kovalerchuk2020SurveyOE}. In comparison, XGNN requires training a new (black box) reinforcement learning agent for each explanation.
\end{itemize} 

\paragraph{Usage of XGNN} We use the implementation of XGNN provided in DIG \cite{dig}. We remark that XGNN is able to generate graphs maximising the prediction score on MUTAG, as shown in \cite{xgnn}. However, we find that XGNN fails to generate a graph with score above 0.8 on IMDB \textit{romance} class, even after thorough tuning of parameters. For the \textit{action} class, we find that XGNN fails to generate a graph beyond a single node. 

\section{Conclusions}

We perform a new explorative study of the presence of concepts in graph neural networks and provide insights which have potential to improve our understanding of GNNs and help practitioners gain a more transparent view of their models. To the best of our knowledge this is the first work that tackles GNN interpretability via neuron-level analysis of GNN representations. Furthermore, we use concepts to provide model-level explanations, which show promising advantages over the current state-of-the-art. This is only an initial step in unveiling the neuron-level concepts that exist in GNNs, and much work can be done in the directions we have outlined in the paper. A promising future direction is to further explore the relationship between neuron interpretability and theoretical issues such as the information bottleneck, over-smoothing and over-squashing. 

\section*{Acknowledgements}
We would like to thank the anonymous reviewers for their insightful feedback and suggestions. We also thank Shuntian Liu, Joe Mayford and Yunqing Xuan for proofreading an earlier manuscript. Han Xuanyuan would like to thank He Yu for her tireless support. 
\bibliography{reference}

\begin{thebibliography}{39}
\providecommand{\natexlab}[1]{#1}

\bibitem[{Amann et~al.(2020)Amann, Blasimme, Vayena, Frey, and
  Madai}]{xaihealthcare}
Amann, J.; Blasimme, A.; Vayena, E.; Frey, D.; and Madai, V.~I. 2020.
\newblock {Explainability for Artificial Intelligence in Healthcare: A
  Multidisciplinary Perspective}.
\newblock \emph{BMC Medical Informatics and Decision Making}, 20(1): 1--9.

\bibitem[{Barbiero et~al.(2022)Barbiero, Ciravegna, Giannini, Li{\`o}, Gori,
  and Melacci}]{lens}
Barbiero, P.; Ciravegna, G.; Giannini, F.; Li{\`o}, P.; Gori, M.; and Melacci,
  S. 2022.
\newblock {Entropy-based logic explanations of neural networks}.
\newblock In \emph{Proceedings of the AAAI Conference on Artificial
  Intelligence}, volume~36, 6046--6054.

\bibitem[{Bau et~al.(2017)Bau, Zhou, Khosla, Oliva, and Torralba}]{netdissect}
Bau, D.; Zhou, B.; Khosla, A.; Oliva, A.; and Torralba, A. 2017.
\newblock {Network Dissection: Quantifying Interpretability of Deep Visual
  Representations}.
\newblock In \emph{Proceedings of the IEEE Conference on Computer Vision and
  Pattern Recognition}, 6541--6549.

\bibitem[{Borgwardt et~al.(2005)Borgwardt, Ong, Schönauer, Vishwanathan,
  Smola, and Kriegel}]{proteins}
Borgwardt, K.~M.; Ong, C.~S.; Schönauer, S.; Vishwanathan, S. V.~N.; Smola,
  A.~J.; and Kriegel, H.-P. 2005.
\newblock {Protein Function Prediction via Graph Kernels}.
\newblock \emph{Bioinformatics}.

\bibitem[{Bussmann et~al.(2021)Bussmann, Giudici, Marinelli, and
  Papenbrock}]{xaiinfinance}
Bussmann, N.; Giudici, P.; Marinelli, D.; and Papenbrock, J. 2021.
\newblock {Explainable Machine Learning in Credit Risk Management}.
\newblock \emph{Computational Economics}, 57(1): 203--216.

\bibitem[{Chen(2017)}]{kde}
Chen, Y.-C. 2017.
\newblock {A Tutorial on Kernel Density Estimation and Recent Advances}.
\newblock \emph{Biostatistics \& Epidemiology}, 1(1): 161--187.

\bibitem[{Dai et~al.(2022)Dai, Zhao, Zhu, Xu, Guo, Liu, Tang, and
  Wang}]{newsurvey}
Dai, E.; Zhao, T.; Zhu, H.; Xu, J.; Guo, Z.; Liu, H.; Tang, J.; and Wang, S.
  2022.
\newblock {A Comprehensive Survey on Trustworthy Graph Neural Networks:
  Privacy, Robustness, Fairness, and Explainability}.
\newblock \emph{arXiv preprint arXiv:2204.08570}.

\bibitem[{Dalvi et~al.(2019)Dalvi, Durrani, Sajjad, Belinkov, Bau, and
  Glass}]{Dalvi2019WhatIO}
Dalvi, F.; Durrani, N.; Sajjad, H.; Belinkov, Y.; Bau, A.; and Glass, J. 2019.
\newblock {What Is One Grain of Sand in the Desert? Analyzing Individual
  Neurons in Deep NLP Models}.
\newblock In \emph{Proceedings of the AAAI Conference on Artificial
  Intelligence}, volume~33, 6309--6317.

\bibitem[{Debnath et~al.(1991)Debnath, Lopez~de Compadre, Debnath, Shusterman,
  and Hansch}]{mutag}
Debnath, A.~K.; Lopez~de Compadre, R.~L.; Debnath, G.; Shusterman, A.~J.; and
  Hansch, C. 1991.
\newblock {Structure-activity relationship of mutagenic aromatic and
  heteroaromatic nitro compounds. correlation with molecular orbital energies
  and hydrophobicity}.
\newblock \emph{Journal of Medicinal Chemistry}, 34(2): 786--797.

\bibitem[{Deeks(2019)}]{xaiinlaw}
Deeks, A. 2019.
\newblock {The Judicial Demand for Explainable Artificial Intelligence}.
\newblock \emph{Columbia Law Review}, 119(7): 1829--1850.

\bibitem[{Georgiev et~al.(2022)Georgiev, Barbiero, Kazhdan,
  Veli{\v{c}}kovi{\'c}, and Li{\`o}}]{acer}
Georgiev, D.; Barbiero, P.; Kazhdan, D.; Veli{\v{c}}kovi{\'c}, P.; and Li{\`o},
  P. 2022.
\newblock {Algorithmic Concept-based Explainable Reasoning}.
\newblock In \emph{Proceedings of the AAAI Conference on Artificial
  Intelligence}, volume~36, 6685--6693.

\bibitem[{Ghorbani et~al.(2019)Ghorbani, Wexler, Zou, and
  Kim}]{Ghorbani2019TowardsAC}
Ghorbani, A.; Wexler, J.; Zou, J.~Y.; and Kim, B. 2019.
\newblock {Towards Automatic Concept-Based Explanations}.
\newblock \emph{Advances in Neural Information Processing Systems}, 32.

\bibitem[{Himmelhuber et~al.(2021)Himmelhuber, Zillner, Grimm, Ringsquandl,
  Joblin, and Runkler}]{himmelhuber2021new}
Himmelhuber, A.; Zillner, S.; Grimm, S.; Ringsquandl, M.; Joblin, M.; and
  Runkler, T.~A. 2021.
\newblock A New Concept for Explaining Graph Neural Networks.
\newblock In \emph{NeSy}, 1--5.

\bibitem[{Kazius, McGuire, and Bursi(2005)}]{toxicophore}
Kazius, J.; McGuire, R.; and Bursi, R. 2005.
\newblock {Derivation and Validation of Toxicophores for Mutagenicity
  Prediction}.
\newblock \emph{Journal of Medicinal Chemistry}, 48(1): 312--320.

\bibitem[{Kim et~al.(2018)Kim, Wattenberg, Gilmer, Cai, Wexler, Viegas
  et~al.}]{Kim2018InterpretabilityBF}
Kim, B.; Wattenberg, M.; Gilmer, J.; Cai, C.; Wexler, J.; Viegas, F.; et~al.
  2018.
\newblock {Interpretability beyond feature attribution: Quantitative testing
  with concept activation vectors (tcav)}.
\newblock In \emph{International Conference on Machine Learning}, 2668--2677.
  PMLR.

\bibitem[{Koh et~al.(2020)Koh, Nguyen, Tang, Mussmann, Pierson, Kim, and
  Liang}]{Koh2020ConceptBM}
Koh, P.~W.; Nguyen, T.; Tang, Y.~S.; Mussmann, S.; Pierson, E.; Kim, B.; and
  Liang, P. 2020.
\newblock {Concept Bottleneck Models}.
\newblock In \emph{International Conference on Machine Learning}, 5338--5348.
  PMLR.

\bibitem[{Kovalerchuk, Ahmad, and Teredesai(2021)}]{Kovalerchuk2020SurveyOE}
Kovalerchuk, B.; Ahmad, M.~A.; and Teredesai, A. 2021.
\newblock {Survey of Explainable Machine Learning with Visual and Granular
  Methods Beyond Quasi-Explanations}.
\newblock \emph{Interpretable Artificial Intelligence: A Perspective of
  Granular Computing}, 217--267.

\bibitem[{Li, Monroe, and Jurafsky(2016)}]{li2016understanding}
Li, J.; Monroe, W.; and Jurafsky, D. 2016.
\newblock {Understanding Neural Networks Through Representation Erasure}.
\newblock \emph{arXiv preprint arXiv:1612.08220}.

\bibitem[{Li, Han, and Wu(2018)}]{li2018deeper}
Li, Q.; Han, Z.; and Wu, X.-M. 2018.
\newblock {Deeper Insights into Graph Convolutional Networks for
  Semi-Supervised Learning}.
\newblock In \emph{Thirty-Second AAAI Conference on Artificial Intelligence}.

\bibitem[{Liu et~al.(2021)Liu, Luo, Wang, Xie, Yuan, Gui, Yu, Xu, Zhang, Liu
  et~al.}]{dig}
Liu, M.; Luo, Y.; Wang, L.; Xie, Y.; Yuan, H.; Gui, S.; Yu, H.; Xu, Z.; Zhang,
  J.; Liu, Y.; et~al. 2021.
\newblock {DIG: A Turnkey Library for Diving into Graph Deep Learning
  Research}.
\newblock \emph{Journal of Machine Learning Research}, 22(240): 1--9.

\bibitem[{Lundberg et~al.(2020)Lundberg, Erion, Chen, DeGrave, Prutkin, Nair,
  Katz, Himmelfarb, Bansal, and Lee}]{lundberg2020local}
Lundberg, S.~M.; Erion, G.; Chen, H.; DeGrave, A.; Prutkin, J.~M.; Nair, B.;
  Katz, R.; Himmelfarb, J.; Bansal, N.; and Lee, S.-I. 2020.
\newblock {From Local Explanations to Global Understanding with Explainable AI
  for Trees}.
\newblock \emph{Nature Machine Intelligence}, 2(1): 56--67.

\bibitem[{Luo et~al.(2020)Luo, Cheng, Xu, Yu, Zong, Chen, and
  Zhang}]{pgexplainer}
Luo, D.; Cheng, W.; Xu, D.; Yu, W.; Zong, B.; Chen, H.; and Zhang, X. 2020.
\newblock {Parameterized Explainer for Graph Neural Network}.
\newblock \emph{Advances in Neural Information Processing Systems}, 33.

\bibitem[{Magister et~al.(2021)Magister, Kazhdan, Singh, and
  Li{\`o}}]{gcexplainer}
Magister, L.~C.; Kazhdan, D.; Singh, V.; and Li{\`o}, P. 2021.
\newblock {GCExplainer: Human-in-the-Loop Concept-based Explanations for Graph
  Neural Networks}.
\newblock \emph{arXiv preprint arXiv:2107.11889}.

\bibitem[{Mu and Andreas(2020)}]{muandreas}
Mu, J.; and Andreas, J. 2020.
\newblock {Compositional Explanations of Neurons}.
\newblock \emph{Advances in Neural Information Processing Systems}, 33:
  17153--17163.

\bibitem[{Samek et~al.(2019)Samek, Montavon, Vedaldi, Hansen, and
  M{\"u}ller}]{samek2019explainable}
Samek, W.; Montavon, G.; Vedaldi, A.; Hansen, L.~K.; and M{\"u}ller, K.-R.
  2019.
\newblock \emph{{Explainable AI: Interpreting, Explaining and Visualizing Deep
  Learning}}, volume 11700.
\newblock Springer Nature.

\bibitem[{Sanchez-Gonzalez et~al.(2020)Sanchez-Gonzalez, Godwin, Pfaff, Ying,
  Leskovec, and Battaglia}]{SanchezGonzalez2020LearningTS}
Sanchez-Gonzalez, A.; Godwin, J.; Pfaff, T.; Ying, R.; Leskovec, J.; and
  Battaglia, P. 2020.
\newblock {Learning to Simulate Complex Physics with Graph Networks}.
\newblock In \emph{International Conference on Machine Learning}, 8459--8468.
  PMLR.

\bibitem[{Selvaraju et~al.(2019)Selvaraju, Das, Vedantam, Cogswell, Parikh, and
  Batra}]{gradcam}
Selvaraju, R.~R.; Das, A.; Vedantam, R.; Cogswell, M.; Parikh, D.; and Batra,
  D. 2019.
\newblock {Grad-CAM: Visual Explanations from Deep Networks via Gradient-Based
  Localization}.
\newblock \emph{International Journal of Computer Vision}, 128: 336--359.

\bibitem[{Shwartz-Ziv and Tishby(2017)}]{shwartz2017opening}
Shwartz-Ziv, R.; and Tishby, N. 2017.
\newblock {Opening the Black Box of Deep Neural Networks via Information}.
\newblock \emph{arXiv preprint arXiv:1703.00810}.

\bibitem[{Simkute et~al.(2021)Simkute, Luger, Jones, Evans, and
  Jones}]{SIMKUTE2021100017}
Simkute, A.; Luger, E.; Jones, B.; Evans, M.; and Jones, R. 2021.
\newblock {Explainability for Experts: A Design Framework for Making Algorithms
  Supporting Expert Decisions More Explainable}.
\newblock \emph{Journal of Responsible Technology}, 7-8: 100017.

\bibitem[{Veli{\v{c}}kovi{\'c}(2022)}]{messagepassing}
Veli{\v{c}}kovi{\'c}, P. 2022.
\newblock {Message Passing All the Way Up}.
\newblock \emph{arXiv preprint arXiv:2202.11097}.

\bibitem[{Vu and Thai(2020)}]{Vu2020PGMExplainerPG}
Vu, M.; and Thai, M.~T. 2020.
\newblock {PGM-Explainer: Probabilistic Graphical Model Explanations for Graph
  Neural Networks}.
\newblock \emph{Advances in Neural Information Processing Systems}, 33:
  12225--12235.

\bibitem[{Welling and Kipf(2016)}]{gcn}
Welling, M.; and Kipf, T.~N. 2016.
\newblock {Semi-Supervised Classification with Graph Convolutional Networks}.
\newblock In \emph{J. International Conference on Learning Representations
  (ICLR 2017)}.

\bibitem[{Wieder et~al.(2020)Wieder, Kohlbacher, Kuenemann, Garon, Ducrot,
  Seidel, and Langer}]{WIEDER20201}
Wieder, O.; Kohlbacher, S.; Kuenemann, M.; Garon, A.; Ducrot, P.; Seidel, T.;
  and Langer, T. 2020.
\newblock {A Compact Review of Molecular Property Prediction with Graph Neural
  Networks}.
\newblock \emph{Drug Discovery Today: Technologies}, 37: 1--12.

\bibitem[{Wu et~al.(2021)Wu, Chen, Shen, Guo, Gao, Li, Pei, and
  Long}]{Wu2021GraphNN}
Wu, L.; Chen, Y.; Shen, K.; Guo, X.; Gao, H.; Li, S.; Pei, J.; and Long, B.
  2021.
\newblock {Graph Neural Networks for Natural Language Processing: A Survey}.
\newblock \emph{arXiv preprint arXiv:2106.06090}.

\bibitem[{Xu et~al.(2019)Xu, Hu, Leskovec, and Jegelka}]{gin}
Xu, K.; Hu, W.; Leskovec, J.; and Jegelka, S. 2019.
\newblock {How Powerful are Graph Neural Networks?}
\newblock \emph{ArXiv}, abs/1810.00826.

\bibitem[{Yanardag and Vishwanathan(2015)}]{deepgraphkernels}
Yanardag, P.; and Vishwanathan, S. 2015.
\newblock {Deep Graph Kernels}.
\newblock In \emph{Proceedings of the 21th ACM SIGKDD International Conference
  on Knowledge Discovery and Data Mining}, 1365--1374.

\bibitem[{Ying et~al.(2019)Ying, Bourgeois, You, Zitnik, and
  Leskovec}]{gnnexplainer}
Ying, Z.; Bourgeois, D.; You, J.; Zitnik, M.; and Leskovec, J. 2019.
\newblock {GNNExplainer: Generating Explanations for Graph Neural Networks}.
\newblock \emph{Advances in Neural Information Processing Systems}, 32.

\bibitem[{Yuan et~al.(2020)Yuan, Tang, Hu, and Ji}]{xgnn}
Yuan, H.; Tang, J.; Hu, X.; and Ji, S. 2020.
\newblock {XGNN: Towards Model-Level Explanations of Graph Neural Networks}.
\newblock \emph{Proceedings of the 26th ACM SIGKDD International Conference on
  Knowledge Discovery \& Data Mining}.

\bibitem[{Zhou et~al.(2016)Zhou, Khosla, Lapedriza, Oliva, and Torralba}]{cam}
Zhou, B.; Khosla, A.; Lapedriza, A.; Oliva, A.; and Torralba, A. 2016.
\newblock {Learning Deep Features for Discriminative Localization}.
\newblock In \emph{Proceedings of the IEEE Conference on Computer Vision and
  Pattern Recognition}, 2921--2929.

\end{thebibliography}

\clearpage
\newpage
\appendix
\onecolumn

\section{Concept extraction}
\label{appendix:conceptextraction}

At each iteration $i$ of the beam search, we maintain a set of concepts of length $i+1$ that are within the beam and combine them with all unary concepts in $\mathcal{C} \cup \{\neg C \mid C \in \mathcal{C}\}$ to obtain a set of candidate concepts $\mathcal{C}_\text{cand}$ of length $i+2$. We then compute the expected divergence score for all concepts in $\mathcal{C}_\text{cand}$ and prune all but the top $T$ concepts. Since we only include the operators of disjunction and conjunction in our logical forms, unary concepts that are less desirable are unlikely to be part of the final formula. This motivates including the negated unary concepts in the set of base concepts. In our implementation a beam width of 10 is used for all models.

\algnewcommand{\IIf}[1]{\State\algorithmicif\ #1\ \algorithmicthen}
\algnewcommand{\EndIIf}{\unskip\ \algorithmicend\ \algorithmicif}

\begin{algorithm}
\caption{GNN concept extraction framework}\label{extract}
\hspace*{\algorithmicindent} \textbf{Input} Model $\mathcal{M}$, training set $\mathcal{D}_\text{train}$, atomic concepts $\mathcal{A}$, depth $l$, beam width $w$ \\
\hspace*{\algorithmicindent} \textbf{Output} Neuron concept scores $\verb|map| : \text{neurons}(\mathcal{M}) \times \mathcal{C} \rightarrow \mathbb{R}$
\begin{algorithmic}[1]
\Procedure{Update }{}
\State $G_\text{batch}, I_\text{batch} \gets \text{batch}(\mathcal{D}_\text{train})$
\State $\mathbf{A} \gets \mathcal{M}_\text{acts}(G)$
\For {$u \in \{1, \ldots, |\text{neurons}(\mathcal{M})| \}$}
\State $\mathbf{s} \gets \textsc{Compare}(\mathbf{A}_{\cdot, u}, I_\text{batch}, \mathcal{T}_u)$ \Comment{get scores for each concept}
\State $\mathcal{T}_u \gets   \text{prune}_w(\mathcal{T}_u, \mathbf{s})$ \Comment{only keep top $w$ concepts}
\State update \verb|map| with $\mathcal{T}_u$ and $\mathbf{s}$
\EndFor
\EndProcedure

\Procedure{Expand}{}
\For {$u \in \{1, \ldots, |\text{neurons}(\mathcal{M})| \}$}
\For {$c \in \mathcal{T}_u$}
    $\mathcal{T}_u \gets \{ C_1 \oplus C_2 \mid C_1, C_2 \in \mathcal{T} \wedge \oplus \in \{ \verb|and|, \verb|or| \} \}$
\EndFor
\EndFor

\EndProcedure

\Procedure{Extract}{}
\State \text{Initialise} $\mathcal{C} \gets \mathcal{A}$, \verb|map| $\gets$ \verb|empty map|
\State Initialise $\mathcal{T}_u \gets \{\} \ \forall u \in \{1, \ldots, |\text{neurons}(\mathcal{M})| \}$
\For {$i \in \{ 1, \ldots, l \}$}
\State \textsc{Update}()
\IIf {$i < l$} \textsc{Expand}()\EndIIf
\EndFor
\EndProcedure

\State Call \textsc{Extract}
\end{algorithmic}
\label{fullalgorithm}
\end{algorithm}

In practice, we can use a GPU to make the search feasible. A vectorised version of the divergence function utilising scatter operations is used. For a given neuron, let $\mathbf{h}$ be the vector of activations across all nodes in $\mathcal{D}$ for that neuron, and let $I$ be the index vector specifying the graph index of each node. Using scatter operations we can create a function $\textsc{Compare}(\mathbf{h}, I, \mathcal{T})$ which computes the divergence scores between the neuron and all concepts in a set $\mathcal{T}$. The exact details are given below:

Given a concept $C$, we can concatenate the concept masks for all graphs and likewise the activations as follows:
\begin{equation}
    \mathbf{a}=\concat_{G \in \mathcal{D}} C(G), \ \ \ \ \ \mathbf{b}=\concat_{G \in \mathcal{D}} \mathbf{X}_G^{(l)}[k,:].
\end{equation}
Both $\mathbf{a}$ and $\mathbf{b}$ have $N$ elements where $N$ is the total number of nodes in all graphs in $\mathcal{D}$. An index vector $I$ can be computed. Let $\sigma$ be a scatter addition operation such that $\sigma(\mathbf{x})=\mathbf{y}$ where $y_i=\sum_{j \mid I_j=i} x_j $. The divergence score between $C$ and all graph activations can be computed as 
\begin{equation}
\overrightarrow{\text{Div}}(\mathbf{a},\mathbf{b},\tau, I)=- \sigma \left (\mathbf{a} \cap \tau(\mathbf{b}) \right  ) \odot_\div \sigma ({\mathbf{a} \cup \tau(\mathbf{b})}) \odot_\times \sigma (\mathbf{a} \odot_\times \tau(\mathbf{b})) ) \odot_\div \sigma(\mathbf{b}),
\end{equation}
where $\odot_\div$ and $\odot_\times$ are element-wise division and multiplication. To simultaneously compute all pairwise divergences between all graphs in $\mathcal{D}$ and all concepts in $\mathcal{C}$, we can stack the $\mathbf{a}$'s for each concept into a matrix $\mathbf{A}$, and also the $\mathbf{b}$'s into a matrix $\mathbf{B}$ and perform the same computation, yielding a function $\overrightarrow{\text{Div}}(\mathbf{A},\mathbf{B},\tau, I) $ that computes a matrix $\mathbf{C} \in \mathbb{R}^{n \times |\mathcal{D}|}$ where the element at position $(i,j)$ indicates the divergence between concept $i$ and the activations in graph $j$.  

This can be then used to compute the concept scores of all concepts. The exact procedure is shown in Algorithm \ref{vectorised}.

\begin{algorithm}
\caption{Vectorised concept scoring}
\hspace*{\algorithmicindent} \textbf{Input} Activations $\mathbf{h} \in \mathbb{R}^{N}$, indices $I$, concepts $\mathcal{T}$ \\
\hspace*{\algorithmicindent} \textbf{Output} $\mathbf{s} \in \mathbb{R}^{|\mathcal{T}|}$
\begin{algorithmic}[1]
\Procedure{Compare}{$\mathbf{h}, I, \mathcal{T}$}
\State $\mathbf{H} \gets$ stack $|\mathcal{T}|$ rows of $\mathbf{h}$ 
\State $\mathbf{s} \gets [-\infty, \ldots, -\infty]$
\For {$\tau \in \text{thresholds}$}
\State $\mathbf{C} \gets \overrightarrow{\text{Div}}(\mathcal{T},\mathbf{H},\tau, I)$
\State $\mathbf{s}^\prime \gets \text{mean}(\mathbf{C})$ \Comment{compute the mean of each row}
\State $\mathbf{s} \gets \mathbf{s} \odot_\text{max} \mathbf{s}^\prime$
\EndFor
\State Return $\mathbf{s}$
\EndProcedure
\end{algorithmic}
\label{vectorised}
\end{algorithm}

\section{Entropy score derivation}
\label{appendix:entropy}

Given neurons $n_i, \ldots, n_k$, our objective is to find a subset $S$ which counterfactually explains the prediction of the model $\hat{y}=\mathcal{M}(G)$. That is, removing the neurons in $S$ from the model should change the prediction $\hat{y}$. By removing a neuron $i$ we refer to the action of setting all $n_i(v):=0$ for all nodes $v$. Let $Y$ be the class probability distribution produced by the model, and let $\mathcal{X}$ be the set of neurons being used for the downstream prediction. We use the notion of mutual information and adopt the following objective:
\begin{equation}
    \argmax_{S} \text{MI}(Y,\mathcal{X}) = \argmax_{S} \Big ( H(Y) - H(Y \mid \mathcal{X}=S) \Big).
\end{equation}
Note that $H(Y)$ is the entropy of the probability distribution outputted for the original input (i.e. when $S$ contains all neurons). Since $Y$ is constant, this is equivalent to $\argmin_{S} H(Y \mid \mathcal{X}=S)$. Hence the objective becomes
\begin{equation}
    \argmin_{S} \ - \frac{1}{r}\sum_{c=1}^r \log P_\mathcal{M}(Y=y_c \mid \mathcal{X}=S),
\end{equation}
which is intractible since the number of possible neuron subsets $S$ is exponential in the number of neurons $k$. Following \citet{gnnexplainer} we interpret $S$ probabilistically and consider a multivariate Bernoulli distribution $\mathcal{S}$ from which $S$ is sampled, such that $\text{Pr}_{\mathcal{S}}(S)=\prod_{i=1}^k \eta_i$, where we use a continuous relaxation and consider $S$ to be a \textit{neuron mask} $\eta \in [0, 1]^k$. The objective becomes
\begin{equation}
    \argmin_{\mathcal{S}} \ - \mathbb{E}_{S \in \mathcal{S}} \left [ \sum_{c=1}^r \log P_\mathcal{M}(Y=y_c \mid \mathcal{X}=S) \right ].
\end{equation}
Using Jensen's inequality we obtain 
\begin{equation}
    \argmin_{\mathcal{S}} \ - \sum_{c=1}^r \log P_\mathcal{M}(Y=y_c \mid \mathcal{X}=\mathbb{E}_{S \in \mathcal{S}} [S]),
\end{equation}
which gives us the following optimisable objective for finding $\eta$:
\begin{equation}
    \argmin_{\mathcal{S}} \ - \sum_{c=1}^r \log P_\mathcal{M}(Y=y_c \mid \mathcal{X}=\sigma(\eta)),
\end{equation}
in which $\sigma$ is the sigmoid function. Note that the convexity assumption of Jensen's inequality is not guaranteed - however, similar to \textsc{GNNExplainer} we show that experimentally good local optima can still be reached. To optimise for a certain class prediction $c^\prime$ rather than the overall prediction, we can find
\begin{equation}
    \argmin_{\mathcal{S}} \ - \log P_\Phi(Y=y_{c^\prime} \mid \mathcal{X}=\sigma(\eta)).
\end{equation}
After the mask $\eta$ is learned, we take $S=\text{round}(\sigma(\eta))$ as the final set of neurons to explain the prediction, where $\text{round}(\cdot)$ performs element-wise rounding to the closest integer. 

\section{Black-box model architectures and training}
\label{appendix:models}

All GNN models we try to explain consist of a stack of $N$ GNN layers (with latent representation size $D$), followed by a global pooling operation and a single fully connected layer to produce the output logits. Models are trained using the Adam optimiser with an L2 weight decay. The exact details of the architecture and training parameters for each model are shown in Table \ref{exactarchitecture}. All models are trained using cross-entropy loss. Since the nature of this work is not to maximise performance, we did not thoroughly tune the hyperparameters. They were predominantly chosen to ensure sufficient performance without leading to overfitting. 

All models are trained on a machine with an RTX 3080 GPU and i7 8700k CPU and 32GB of memory. This setup was sufficient for performing all experiments.

\begin{table}[!htb]
\centering
\begin{tabular}{l|cccc}
                    & MUTAG         & PROT.      & IMDB        & REDD.     \\ \cline{1-5}
$N$   & 3 &  $2$  & $3$ & $3$ \\                
$D$   & $64$ &  $64$  & $64$ & $64$ \\
Global pooling & Add & Add & Add & Add        \\
Layer type     & GIN & GIN & GIN & GCN    \\
Output logits        & 2        & 2 &2 & 2        \\ 
Learning rate & 10e-4 & 10e-4 & 4e-3 & 10e-4 \\
Epochs & 850 & 700 & 1000 & 20000 \\
L2 decay & 10e-4 & 10e-4 & 10e-4 & 10e-4 \\ 
Early stop & N/A & 60 & 400 & 1000  \\
Batch size & 32 & 32 & 32 & 32
\end{tabular}
\caption[Exact details of the GNN models used to produce our experimental results.]{Exact details of the GNN models used to produce our experimental results. $D$ is the latent dimension size, and $N$ is the number of GNN layers.}
\label{exactarchitecture}
\end{table}
\newpage
\section{Investigating concepts vs depth}
\label{appendix:depth}

\begin{figure}[!htb]
\centering
\subfloat[\centering MUTAG]{{\includegraphics[height=5cm]{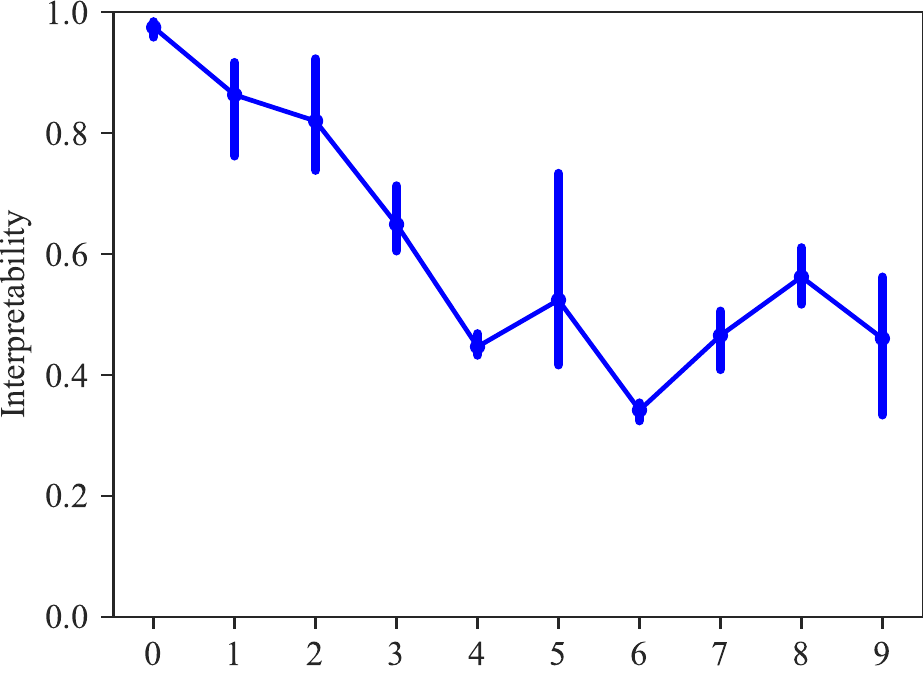} }}%
\subfloat[\centering Reddit-B]{{\includegraphics[height=5cm]{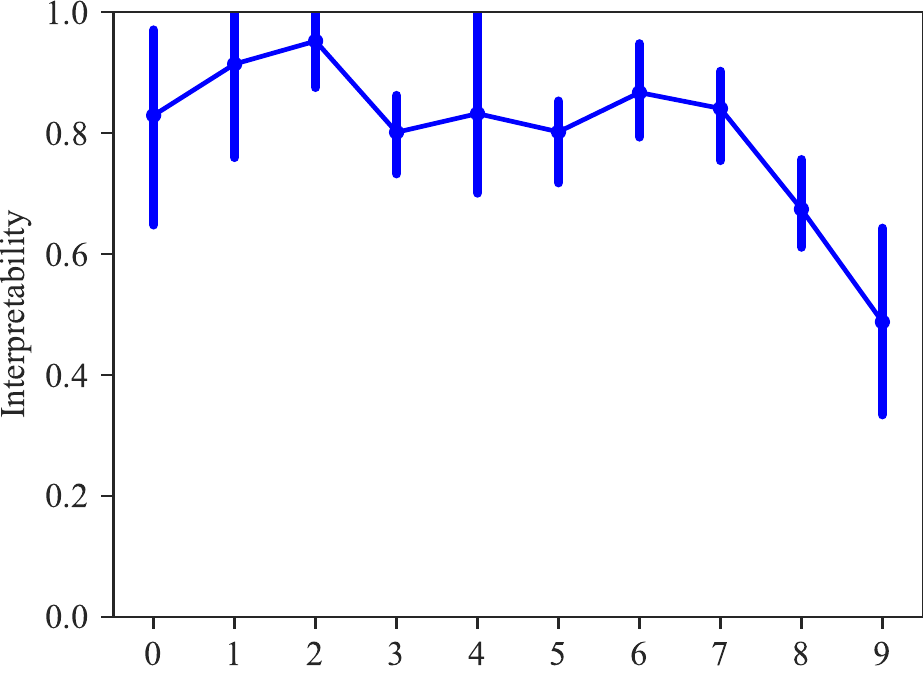} }}%
\caption{The mean neuron interpretability for the 10-layer deep models. Error bars are produced from training three different models with different train-test splits.}
\label{fig:depthexperiment}
\end{figure}

\newpage
\section{Base concepts}
\label{appendix:base}

\begin{table*}[!htb]
\centering
\begin{tabular}{ p{2cm} p{14cm}  }
\hline 
\textbf{Task}  & \textbf{Concepts} \\
 \hline
MUTAG & \text{is}(C) 
\text{is}(N) 
\text{is}(O) 
\text{is}(Cl) 
\text{is}(Br) 
\text{is}(I) 
\text{is}(F) 
\text{next-to}(C) 
\text{next-to}(N) 
\text{next-to}(O) 
\text{next-to}(Cl) 
\text{next-to}(Br) 
\text{next-to}(I) 
\text{next-to}(F) 
\text{next-to}(C,C) 
\text{next-to}(C,C,C) 
\text{next-to}(N,N) 
\text{next-to}(N,N,N) 
\text{next-to}(O) 
\text{next-to}(O,O) 
\text{next-to}(O,O,O) 
\text{degree-is}(1) 
\text{degree-is}(2) 
\text{degree-is}(3)
\text{degree-is}(4)
\text{nb-degree-is}(1)
\text{nb-degree-is}(2)
\text{nb-degree-is}(3) \\
Proteins & \text{is}(A)
\text{is}(B)
\text{is}(C)
\text{next-to}(A)
\text{next-to}(A,A)
\text{next-to}(A,A,A)
\text{next-to}(B)
\text{next-to}(B,B)
\text{next-to}(B,B,B)
\text{next-to}(C)
\text{next-to}(C,C)
\text{next-to}(C,C,C)
\text{next-to}(C,C,C,C)
\text{next-to}(C,C,C,C,C)
\text{nb-next-to}(A)
\text{nb-next-to}(A,A)
\text{nb-next-to}(A,A,A)
\text{nb-next-to}(B)
\text{nb-next-to}(B,B)
\text{nb-next-to}(B,B,B)
\text{nb-next-to}(C)
\text{nb-next-to}(C,C)
\text{nb-next-to}(C,C,C) \\
IMDB-B & $\text{degree-greater}(1), \text{degree-greater}(3), \ldots, \text{degree-greater}(99)$ 
\text{nb-degree-greater}(10,1) 
\text{nb-degree-greater}(10,2)  
\text{nb-degree-greater}(10,3)  
\text{nb-degree-greater}(10,4)  
\text{nb-degree-greater}(10,5)  
\text{nb-degree-greater}(30,1)  
\text{nb-degree-greater}(30,2)  
\text{nb-degree-greater}(30,3)  
\text{nb-degree-greater}(30,4)  
\text{nb-degree-greater}(30,5) \\
Reddit-Binary &  $\text{degree-greater}(1), \text{degree-greater}(4), \ldots, \text{degree-greater}(99)$
\text{nb-degree-greater}(5,1)  
\text{nb-degree-greater}(5,2)
\text{nb-degree-greater}(10,1)  
\text{nb-degree-greater}(10,2)
\text{nb-degree-greater}(30,1)
\text{nb-degree-greater}(30,2)
\text{nb-degree-equal}(1,1) 
\text{nb-degree-equal}(1,2)
\text{nb-degree-equal}(1,3)
\text{nb-degree-equal}(1,4)
\text{nb-degree-equal}(1,10) \\
 \hline
\end{tabular}
\caption{Atomic concepts for the concept extraction process.}
\label{atomicconcepts}
\end{table*}

\begin{table*}[!htb]
\centering
\begin{tabular}{ l p{6cm} p{6cm} }
\hline 
\textbf{Concept}  & \textbf{Description} & \textbf{Displayed form} \\
 \hline
$\text{is}(A)$ & The node is of type $A$ & $\text{is}(A)$ \\
$\text{next-to}(A_1, \ldots, A_n)$ & The neighbourhood of the node contains all types $A_1, \ldots, A_n$ & $\text{next-to}(A_1, \ldots, A_n)$ \\
$\text{nb-next-to}(A_1, \ldots, A_n)$ & The neighbourhood of the node contains a node whose neighbourhood contains all types $A_1, \ldots, A_n$ & $\exists n \in \mathcal{N}.\ \text{next-to}_n(A_1\ldots A_n)$ \\
$\text{degree-is}(X)$ & The degree of the node is exactly $X$  & $\text{degree-is}(X)$  \\
$\text{nb-degree-is}(X)$ & The node has a neighbour whose degree is exactly $X$ &  $\exists n \in \mathcal{N}.\ \text{degree-is}_n(X)$ \\
$\text{nb-degree-equal}(X,Y)$ & The node has at least $X$ neighbours whose degree is exactly $Y$ & $\exists n_1,\ldots,n_X \in \mathcal{N}.$ 

$\ \ \ \ \forall j \in 1, \ldots, X.\ \text{degree-is}_{n_j}(Y)$ \\
$\text{degree-greater}(X)$ & The node's degree is greater than $X$ & $\text{deg-greater}(X)$  \\
$\text{nb-degree-greater}(X,Y)$ & The node has at least $X$ neighbours whose degree is greater than $Y$ & $\exists n_1,\ldots,n_X \in \mathcal{N}.$ 

$\ \ \ \ \forall j \in 1, \ldots, X.\ \text{degree-greater}_{n_j}(Y)$  \\
 \hline
\end{tabular}
\caption{Descriptions of the atomic concepts. $A$ denotes a node type and $X,Y$ denote integers.}
\label{atomicdesc}
\end{table*} 

\newpage


\section{PROTEINS concepts}
\label{appendix:proteins}

\begin{figure*}[!htb]
\centering
\includegraphics[width=.4\textwidth]{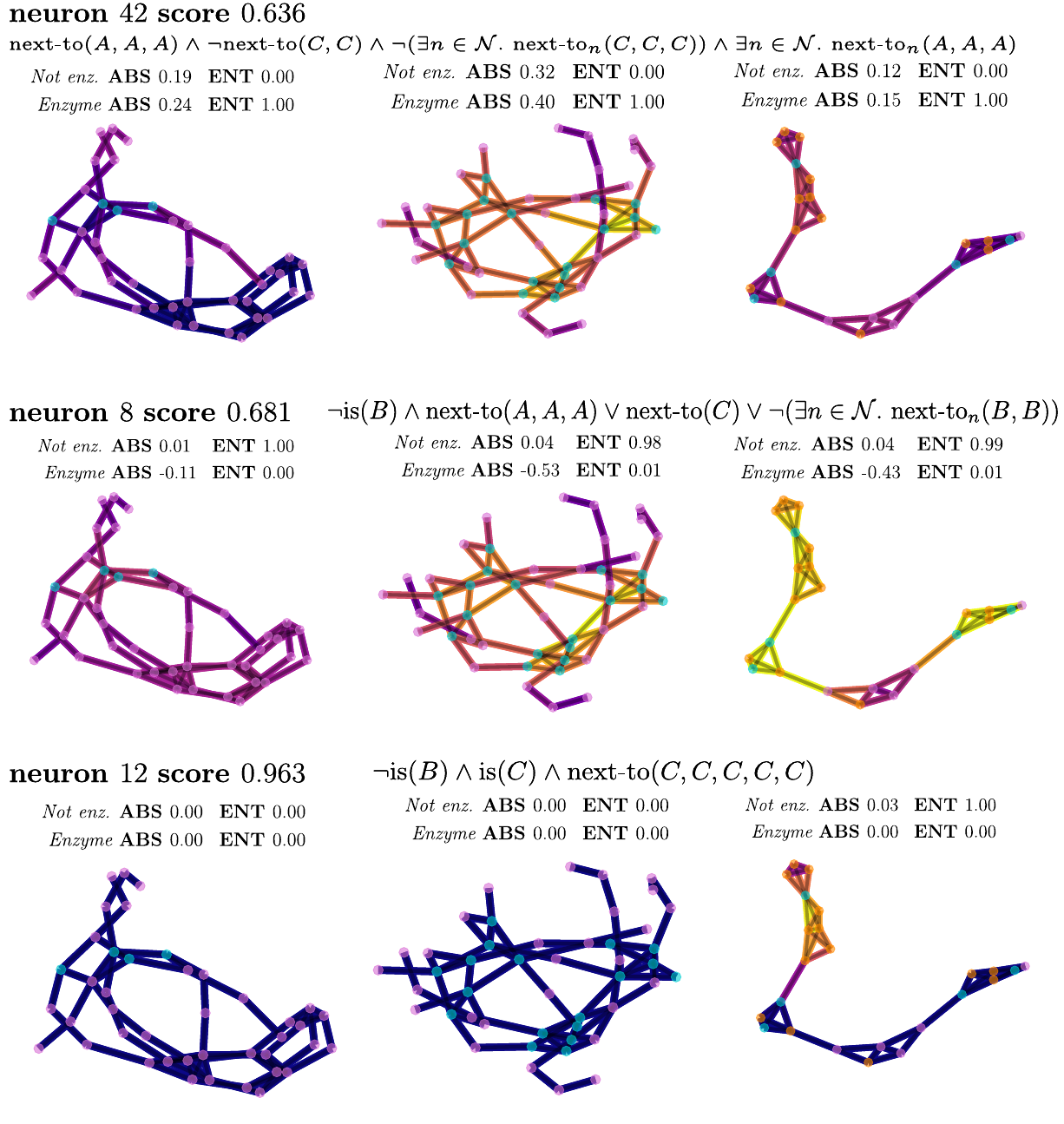}
\caption[Protein concepts]{Protein concepts. Node colours represent amino acid type: types A, B, C are blue, pink and orange respectively.}
\vspace{-1em}
\label{proteinsconcepts}
\end{figure*}

Visualisations of the concepts are provided in Figure \ref{proteinsconcepts}. We discover concepts that are sensitive to the type of amino acid in the immediate neighbourhood of a node: for instance, \textbf{neuron 42} is activated in regions with high concentration of $A$ amino acids, but is surpressed in regions with high concentration of $C$ amino acids. Likewise, \textbf{neuron 12} is only activated in regions with high $C$ concentration.

\section{Additional model-level explanations}
\label{appendix:explanations}

\begin{figure*}[!htb]
\centering
\includegraphics[width=.7\textwidth]{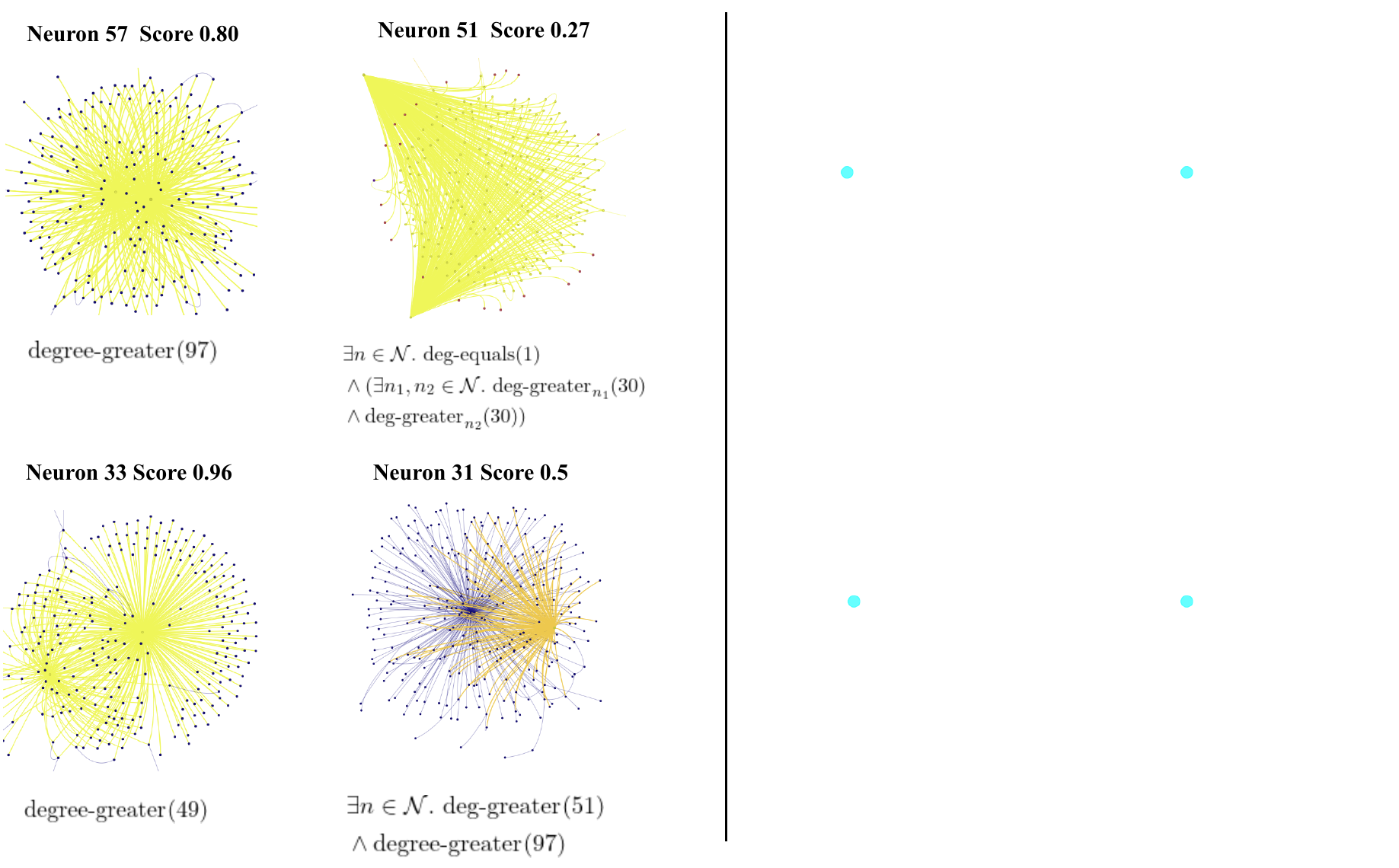}
\caption{Comparison of our approach against XGNN on producing global explanations for the Reddit-Binary Q\&A class. Note that XGNN fails to generate a graph beyond a single node.}
\vspace{-1em}
\label{proteinsconcepts}
\end{figure*}

\end{document}